\documentclass[10pt,twocolumn,letterpaper]{article}

\usepackage{cvpr}              

\usepackage{graphicx}
\usepackage{amsmath}
\usepackage{amssymb}
\usepackage{booktabs}

\usepackage[pagebackref,breaklinks,colorlinks]{hyperref}

\usepackage{multirow}

\usepackage{arydshln}

\newcommand{\cut}[1]{}

\newcommand{\todo}{\textcolor{red}{todo~}}

\newcommand{\red}[1]{\textcolor{red}{#1}}
\newcommand{\blue}[1]{\textcolor{blue}{#1}}

\newcommand{\reid}{re-id}

\newcommand{\name}{\texttt{DeepChange}}

\usepackage[capitalize]{cleveref}
\crefname{section}{Sec.}{Secs.}
\Crefname{section}{Section}{Sections}
\Crefname{table}{Table}{Tables}
\crefname{table}{Tab.}{Tabs.}

\def\cvprPaperID{8791} 
\def\confName{CVPR}
\def\confYear{2022}

\begin{document}

\title{DeepChange: \\A Large Long-Term Person Re-Identification Benchmark with Clothes Change}

\author{Peng Xu\\
University of Oxford \\
{\tt\small peng.xu@eng.ox.ac.uk}
\and
Xiatian Zhu \\
University of Surrey \\
{\tt\small xiatian.zhu@surrey.ac.uk}
}
\maketitle

\begin{abstract}
Existing person re-identification (\reid{}) works
mostly consider {\em short-term} application scenarios without {\em clothes change}.
In real-world, however, we often dress differently across 
space and time.
To solve this contrast, a few recent attempts have been made on long-term \reid{} with clothes change. 
Currently, one of the most significant limitations in this field is the lack of a large realistic benchmark.
In this work, we contribute a large, realistic long-term person re-identification benchmark, named as \name{}. 
It has several unique characteristics:
{\bf (1)} Realistic and rich personal appearance (\eg, clothes and hair style) and variations:
Highly diverse clothes change and styles, with varying reappearing gaps in time from minutes to seasons, different weather conditions (\eg, sunny, cloudy, windy, rainy, snowy, extremely cold) and events (\eg, working, leisure, daily activities).
{\bf (2)} 
Rich camera setups: Raw videos were recorded by $17$ outdoor varying-resolution cameras 
operating in a real-world surveillance system.
%
{\bf (3)}
%
The currently largest number of 
($17$) cameras, ($1,121$) identities, and ($178,407$) bounding boxes,
over the longest time span ($12$ months).
%
Further, {we investigate multimodal fusion strategies for tackling the clothes change challenge.
Extensive experiments show that our fusion models outperform a wide variety of state-of-the-art models on \name{}.
Our dataset and documents are available at \href{https://github.com/PengBoXiangShang/deepchange}{https://github.com/PengBoXiangShang/deepchange}}
\end{abstract}

\begin{figure*}[!t]
	\centering	
	\includegraphics[width=0.9\linewidth]{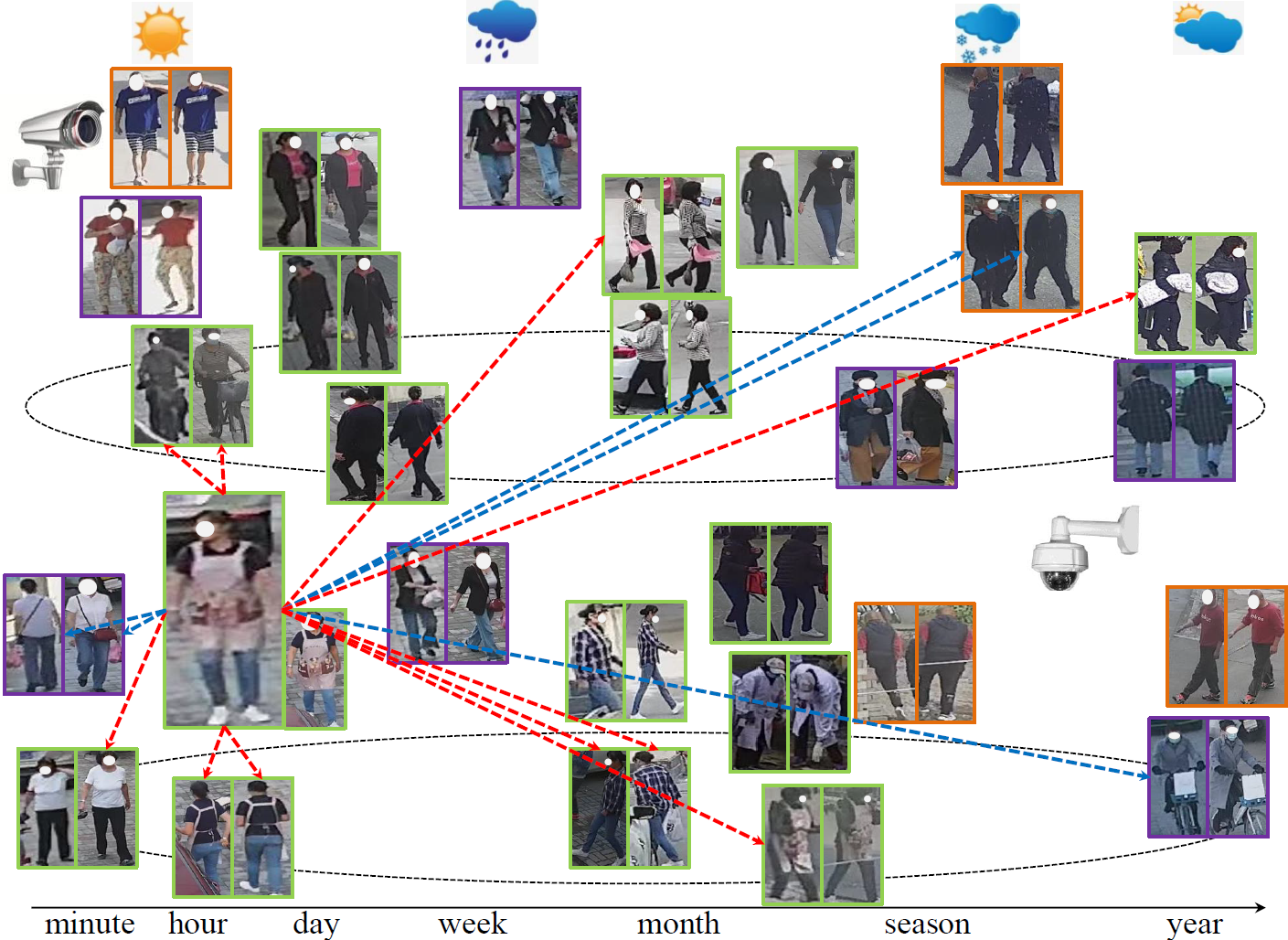}
	\caption{{\bf Motivation}: 
	An average person often changes the clothes 
	over time and space,
	as shown by three random persons (color coded).
	Conventional \reid{} settings usually assume
	stationary clothes per person and are hence valid only for 
	the short-term application scenarios.
	For long-term \reid{} cases, we must consider the unconstrained {\em clothes change} challenge.
	Note, for clarity only part of true (\red{red arrow}) and false (\blue{blue arrow}) matches are plotted.
	%
	Best viewed in color.}
	\label{fig:motivation}
\end{figure*}

\section{Introduction}
\label{sec:introduction}

Person re-identification (\reid{}) aims to match the person identity classes of bounding box images extracted from non-overlapping camera views \cite{gong2014person,zheng2016person,ye2021deep}.
Extensive \reid{} models were developed in the past decade
\cite{gray2008viewpoint,prosser2010person,zheng2012reidentification,zhao2013unsupervised,zhang2016learning,xiao2016learning,zheng2017unlabeled,hermans2017defense,yu2017cross,lan2017deep,chenyanbei2017person,chen2017person,li2018harmonious,sun2018beyond,li2018unsupervised,jiao2018deep,chang2018multi,li2019unsupervised,yu2019unsupervised,ge2020mutual,zhong2020learning,yin2020fine,cheng2020inter,zhu2021intra,zhou2021osnet}, thanks to the availability of reasonably sized benchmarks \cite{wei2018person,zheng2015scalable,li2014deepreid,wang2018transferable}.
The majority of these methods
consider {\em the short-term search scenario with a strong assumption that the appearance (\eg, clothes, hair style) of each person is stationary}. 
We name this conventional setting as
{\em short-term \reid{}} in the follows.
Unfortunately, this assumption would be easily broken once 
the search time span is enlarged to long periods (such as days, weeks, or even months) as an average person often changes the outfit during different day time and across different weathers, daily activities and social events.
As shown in Figure~\ref{fig:motivation},
a specific person was dressed the same only in a short time (\eg, minutes or hours)
but with different clothes/hairs and associations over a long time and across seasons and weathers.
Relying heavily on the clothes appearance, the previous short-term \reid{} models are unsuitable and ineffective 
in dealing with unconstrained clothes changes over space and time.

Recently there have been a few studies for tackling 
the long-term \reid{} situations focusing on clothes change
\cite{yang2019person,huang2019celebrities,qian2020long,li2020learning,wan2020person,wang2020benchmark,wang2014person}.
Since there are no appropriate datasets publicly available,
to enable research these works introduced several small-scale long-term \reid{} datasets
by using web celebrity images \cite{huang2019beyond}, synthesizing pseudo person identities \cite{wan2020person,li2020learning}, 
collecting person images under simulated surveillance settings
\cite{yang2019person,qian2020long,wan2020person,wang2020benchmark}.
While these dataset construction efforts are useful in initiating the research, it is obvious that a real, large-scale long-term \reid{} benchmark is missing and highly demanded.
The impact and significance of such a benchmark in driving the research progress has been repeatedly demonstrated in the short-term \reid{} case \cite{zheng2015scalable,li2014deepreid,wei2018person,wang2014person}. 
However, it is much more challenging to establish a large \reid{} dataset with clothes change as compared to the short-term counterparts. This is due to that:
(1) More video data need to be collected and processed over long time periods;
(2) Labeling person identity becomes much more difficult when a person is dressed up with different outfit from time to time.
%
%
Regardless, we believe that it is worthwhile to overcome all these challenges.

To facilitate the research towards the applications of long-term person search in reality, we contribute the first large-scale person \reid{} dataset with native/natural appearance (mainly clothes) changes, termed as {\name{}}.
Different from existing datasets, \name~has several unique characteristics:
(1) The raw videos are collected in a real-world surveillance camera network deployed at a residential block where rich scenes and realistic background are presented over time and space.
(2) The videos cover a period of 12 months with a wide variety of different weathers and contexts. To the best of our knowledge, this is the longest time duration among all \reid{} datasets, presenting natural/native personal appearance (\eg, clothes and hair style) variations
with people from all walks of life.
(3) Compared to existing long-term \reid{} datasets, 
it contains the largest number of (17) cameras, (1121) identities, and (178K) bounding boxes. 
Overall, \name~is the only realistic, largest long-term person \reid{} dataset, created using the real-world surveillance videos.

We make the following {\bf contributions}:
(1) A large scale long-term person \reid{} dataset, called \name, is introduced. 
Compared with existing alternative datasets, this dataset offers more realistic and more challenging person \reid{} tasks over long time with native appearance changes.
With a much larger quantity of person images for model training,
validation, and testing, \name{} provides a more reliable and indicative test bed for future research.
(2) 
{We conduct extensive experiments
on the \name~dataset, including seminal CNNs
\cite{he2016deep,huang2017densely,sandler2018mobilenetv2,Szegedy_2016_CVPR}, Transformers \cite{dosovitskiy2020image,touvron2021training}, and
state-of-the-art long-term \reid{} models \cite{huang2019beyond}.
%
(3) 
To tackle the clothes change challenge, we investigate multimodal fusion strategies (\eg, gray images, edge maps \cite{ZitnickECCV14edgeBoxes}, key points \cite{cao2019openpose}) and achieve new state-of-the-art results on \name{}.}

\section{DeepChange Benchmark}
\label{sec:dataset}

\begin{figure*}[!t]
	\centering	
	\includegraphics[width=\linewidth]{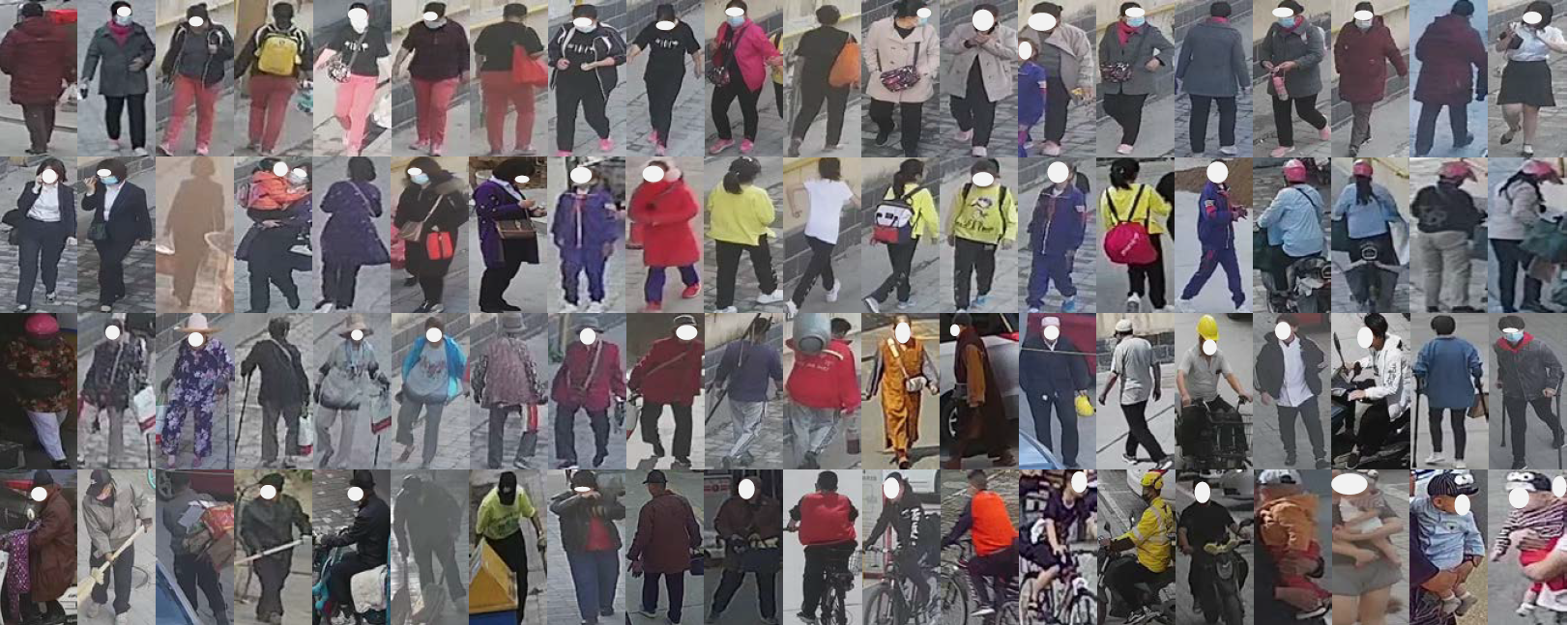}
		
	\caption{Image samples of random identities in \name{}. Identities from top left to bottom right: an aunt (bbox\#1-\#19), an office lady (bbox\#20-\#27), a pupil (bbox\#28-\#36), a newspaper delivery (bbox\#37-\#41), an older aunt (bbox\#42-\#49), a worker (bbox\#50-\#51), a nun (bbox\#52-\#53), a Muslim man (bbox\#54-\#56), a chef (bbox\#57-\#58), a disabled person (bbox\#59-\#60), a dustman (bbox\#61-\#70), a middle school student (bbox\#71-\#74), a food delivery (bbox\#75-\#76), a baby (bbox\#77-\#80). Best viewed in color.}
	\label{fig:identity-diversity}
\end{figure*}

\begin{figure*}[!t]
	\centering	
	\includegraphics[width=\linewidth]{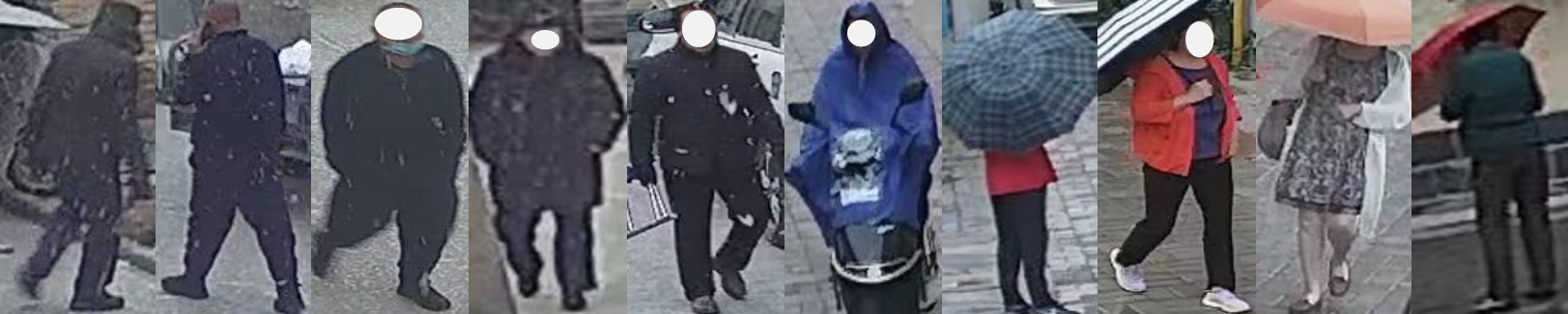}
		
	\caption{Samples collected in snow (bbox\#1-\#5) and rain (bbox\#6-\#10) weather from \name{}. Best viewed in color.}
	\label{fig:snow-and-rain}
\end{figure*}


\noindent{\bf Venue and climate}
Our raw videos were collected from a real-world surveillance system for a wide ($14$ hectares) and dense block, 
with the middle temperate continental monsoon climate.
This venue has various scenes, including crowded streets, shops, restaurants, construction sites, residential buildings, physical exercise areas, car parking, sparsely populated corner, \etc. 
Thus it has two major advantages:
(i) Identity Diversity: Persons cross a wide range of ages and professions, \eg, lactating baby, very older person, office lady, elementary student, high school student, worker, deliveryman, religious. Some identity examples are illustrated in Figure~\ref{fig:identity-diversity}.
(ii) Weather Diversity: During our collecting, we have observed a temperature variation ranging from $-30^{\circ}C$ (in winter) to $35^{\circ}C$ (in summer). 
Therefore, we have collected persons appearing in various weathers, \eg, sunny, cloudy, windy, rainy, snowy, extremely cold. Some image samples with snow are shown in Figure~\ref{fig:snow-and-rain}, where the snow can cause noticeable appearance changes on both clothes or background.
Altogether, the identity and weather diversities will be embodied in drastic appearance changes, enabling realistic long-term \reid{} benchmarking with our \name{} dataset.
Figure
\ref{fig:identity-diversity}
demonstrates some bounding boxes of an identity randomly selected from our dataset, where we can observe dramatic appearance changes across weathers, seasons, \etc.

\vspace{0.1cm}
\noindent{\bf Security camera}
Our raw videos were recorded by $17$ security cameras with a speed of $25$ FPS (frames per second) and different resolutions ($1920\times1080$ spear camera $\times14$, $1280\times960$ spherical camera $\times3$). {These $17$ cameras are part of a large-scale video surveillance system.} In particular, these cameras are mounted on the exterior walls of buildings or on lamp posts, and their height is approximately $3$ to $6$ meters. These cameras are monitoring various views including crowded streets, construction sites, physical exercise areas, car parking, sparsely populated corner, \etc. Therefore, these cameras provide diverse scenes, identities, behaviors, events, \etc.

\begin{figure*}[!t]
	\centering
	\begin{subfigure}{0.31\linewidth}
		\includegraphics[width=\linewidth]{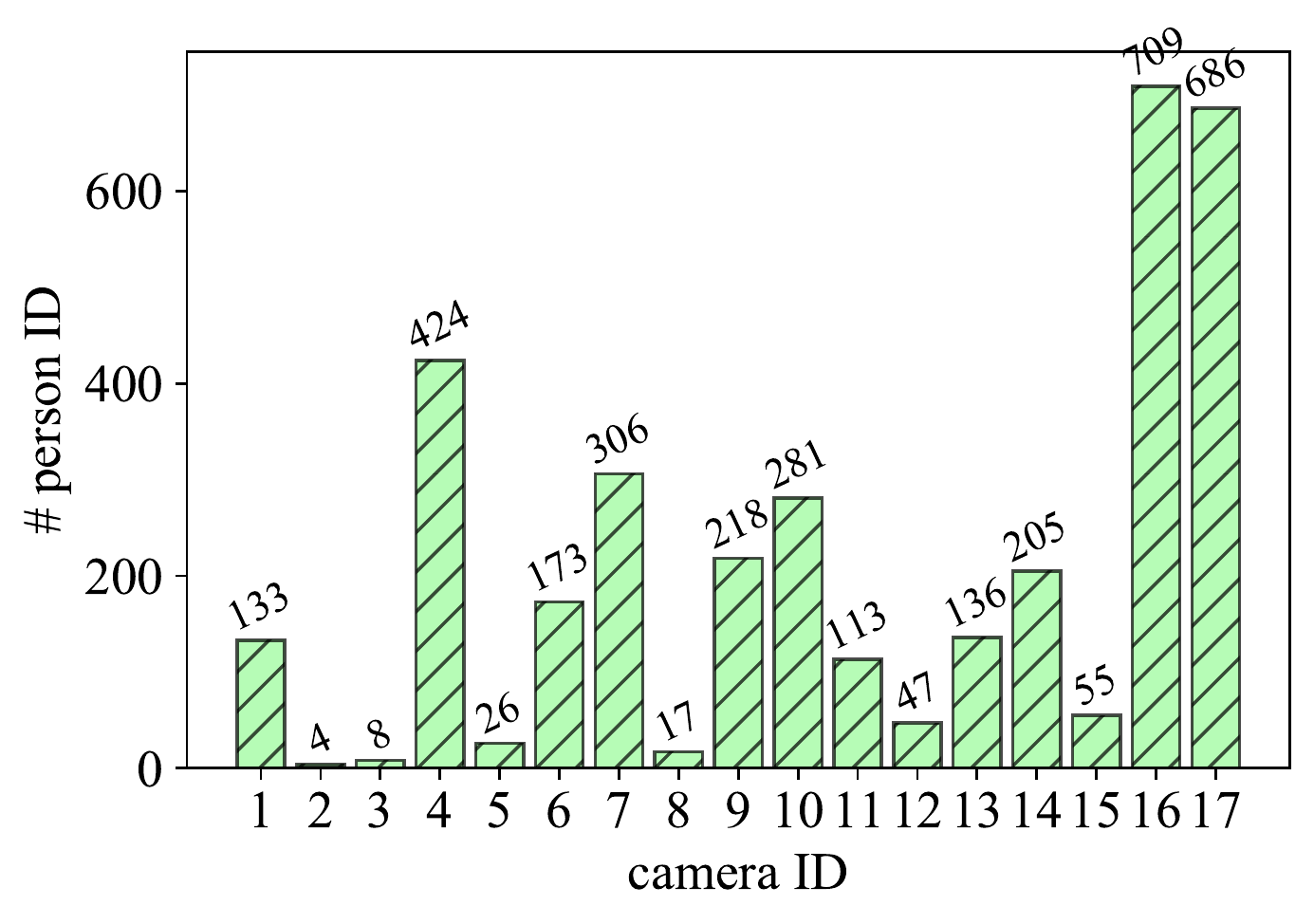}
		\caption{~}
		\label{fig:cid-pid-bar}
	\end{subfigure}
    \begin{subfigure}{0.33\linewidth}
		\includegraphics[width=\linewidth]{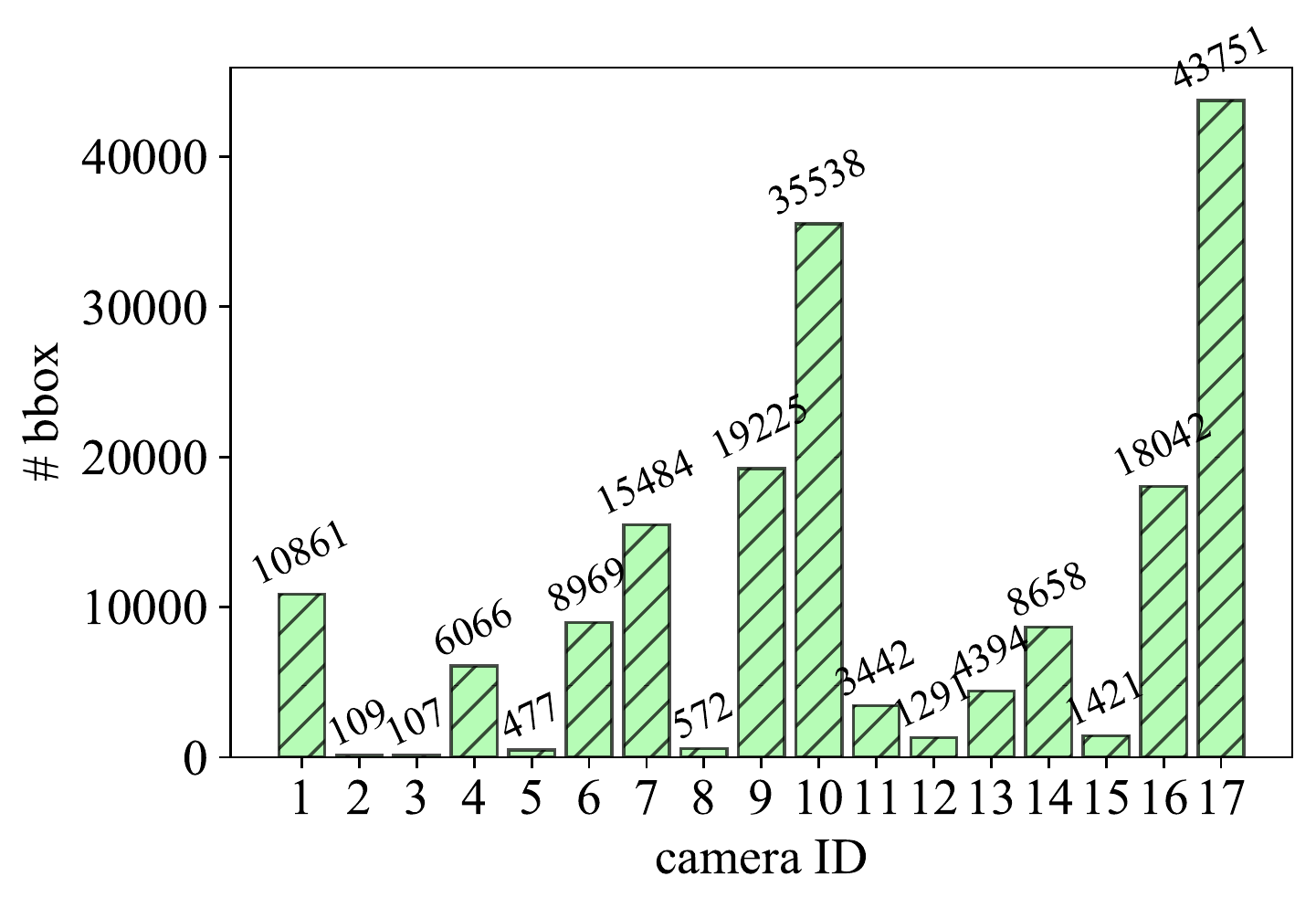}
		\caption{~}
		\label{fig:cid-bbox-bar}
	\end{subfigure}
    \begin{subfigure}{0.32\linewidth}
		\includegraphics[width=\linewidth]{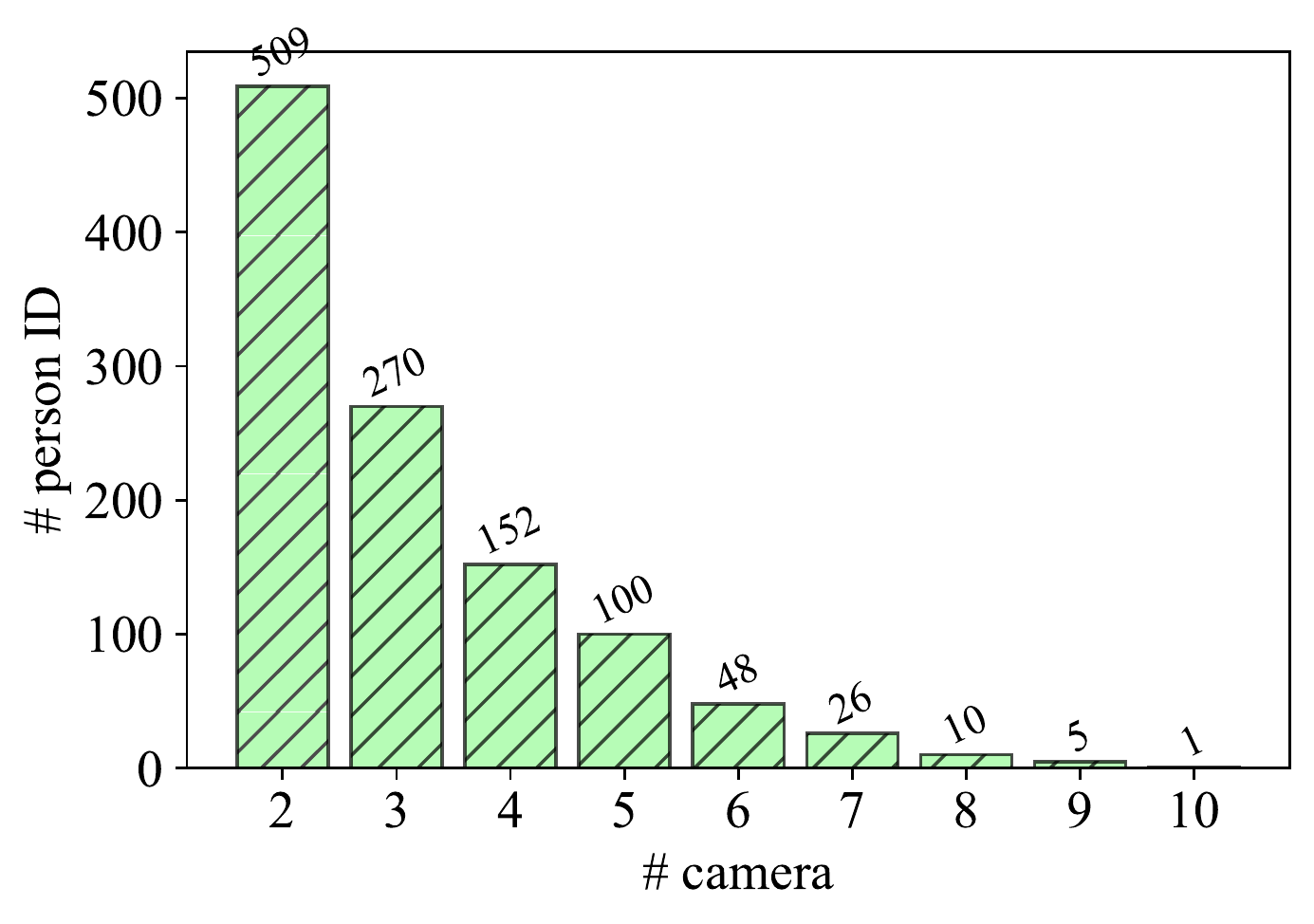}
		\caption{~}
		\label{fig:pid-cid-bar}
	\end{subfigure}
	\begin{subfigure}{0.32\linewidth}
		\includegraphics[width=\linewidth]{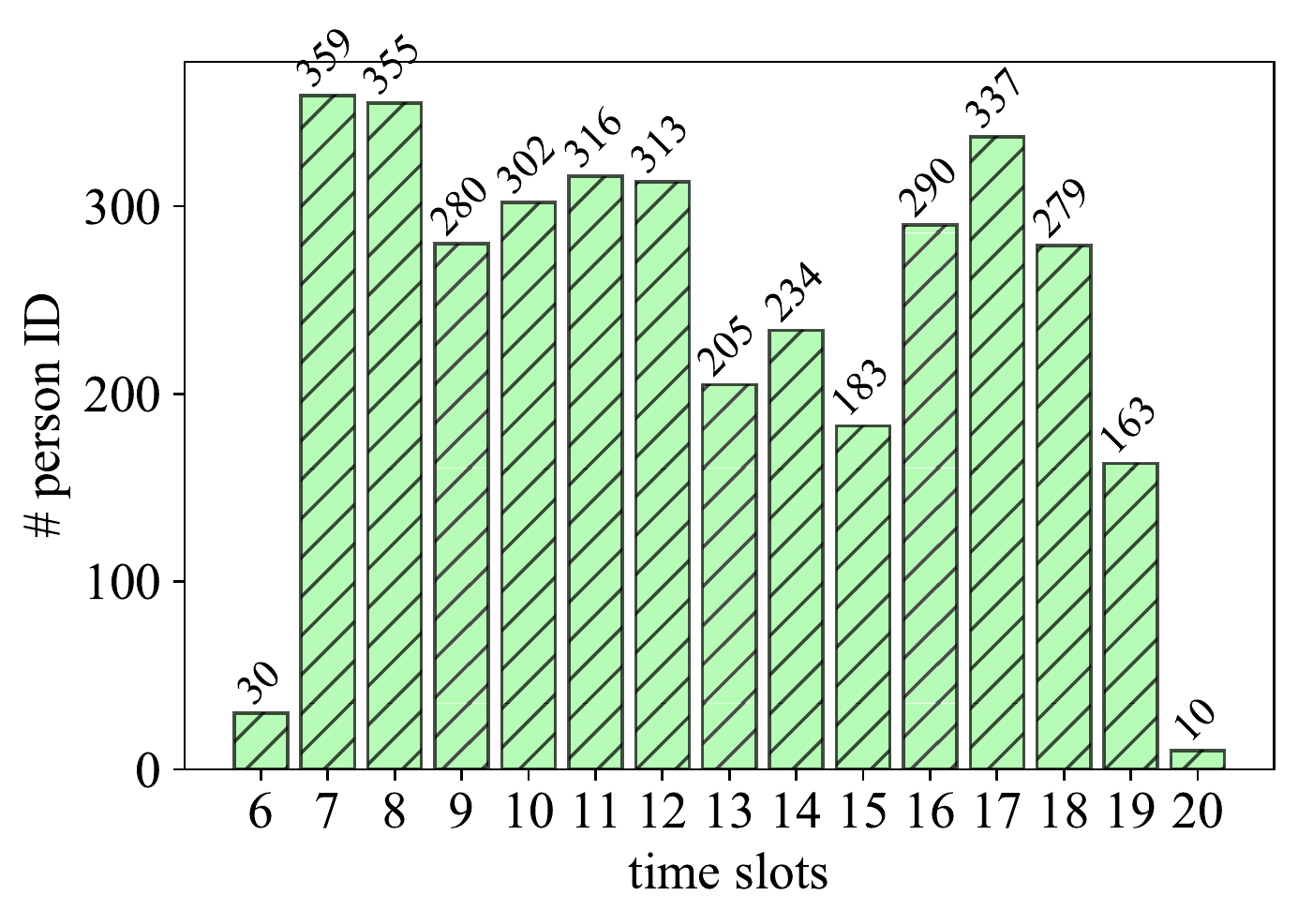}
		\caption{~}
		\label{fig:hour-pid-bar}
	\end{subfigure}
    \begin{subfigure}{0.33\linewidth}
		\includegraphics[width=\linewidth]{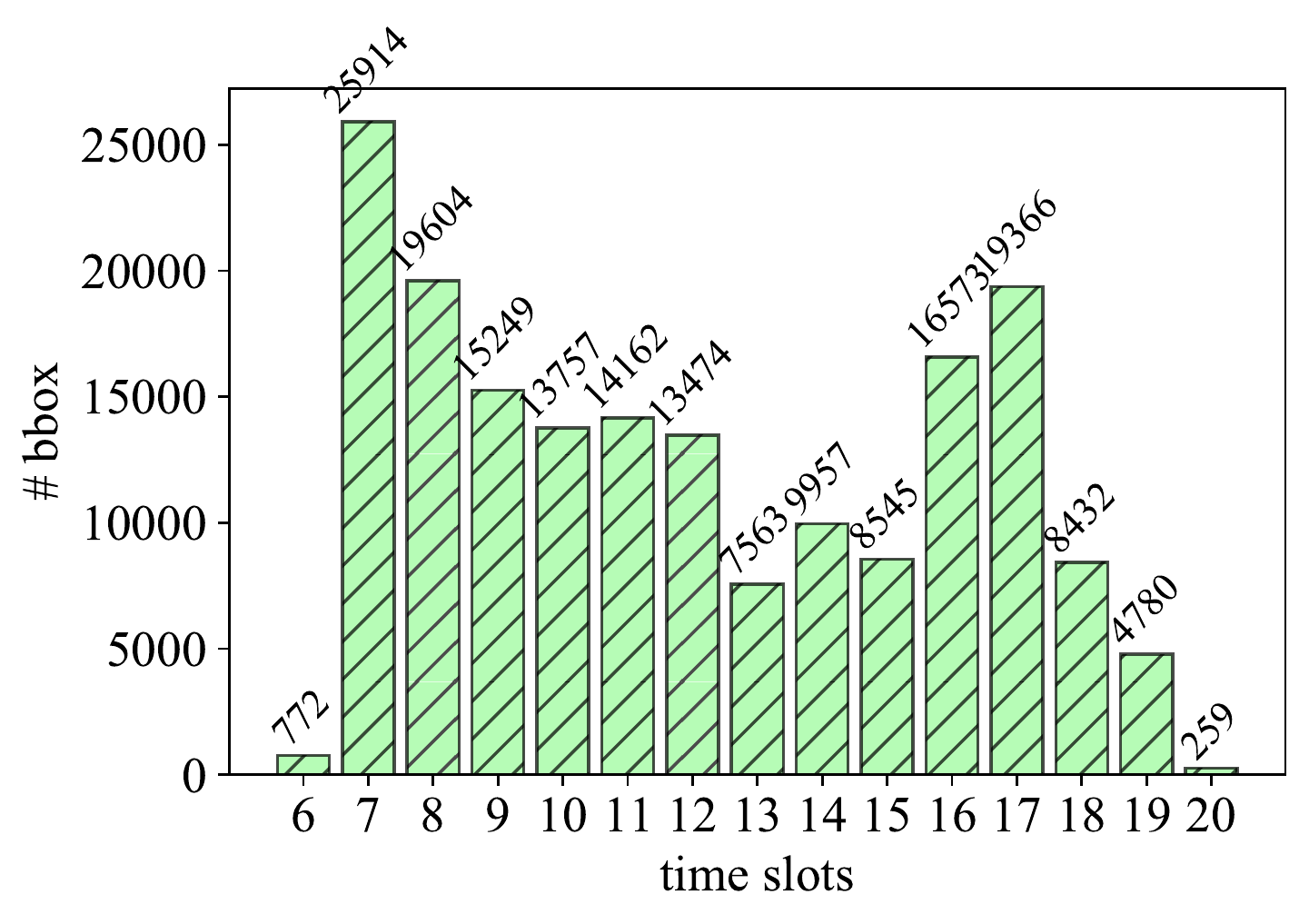}
		\caption{~}
		\label{fig:hour-bbox-bar}
	\end{subfigure}
	\begin{subfigure}{0.30\linewidth}
		\includegraphics[width=\linewidth]{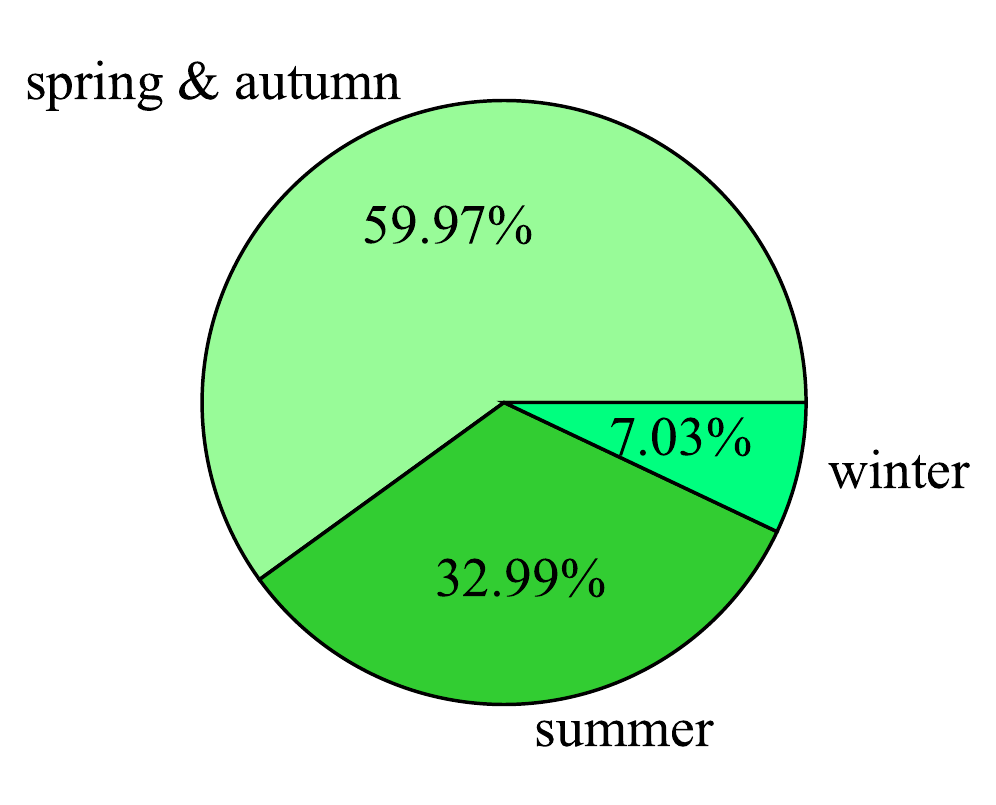}
		\caption{~}
		\label{fig:season-bbox-pie}
	\end{subfigure}
		
	\caption{Statistical analysis of the \name{} dataset. (a): Identity size captured by \#1-\#17 cameras, (b): Bounding box amounts across cameras, (c): Distribution of identity sizes captured by different camera numbers, (d): Identity size in different time slots, (e): Bounding box size in different time slots, (f): Bounding box ratios across seasons (clothes styles).}
	\label{fig:statistic-analysis}
\end{figure*}

\vspace{0.1cm}
\noindent{\bf Video collecting}
Our collecting is a very long course over 12 months across two different calendar years. Every month, we randomly collected raw videos on $7$ to $10$ randomly selected days to cover as much weather as possible. On each selected day, we collected video {from dawn to night} to record comprehensive light variations. A huge amount of videos can provide highly diverse clothes changes and dressing styles, with the reappearing gap in time ranging from minutes, hours, and days to weeks, months, seasons, and years. The permission of using these videos was granted by the
owner/authority for research purposes only.

\begin{table}[!t]
  \caption{Dataset splitting of \name.}
  \label{dataset-split-table}
  \centering
  \resizebox{\linewidth}{!}{
  \begin{tabular}{l l l l}
    \hline

 & Train set & Validation set & Test set \\

    \hline
    \hline
\# Person & 450 & 150 & 521 \\    
\multirow{2}{*}{\# Box} & \multirow{2}{*}{75,083} & Probe: 4,976 & Probe: 17,527 \\
 &  & Gallery: 17,865 & Gallery: 62,956 \\

    \hline
  \end{tabular}
  }
\end{table}

\vspace{0.1cm}
\noindent{\bf Annotation }
It is much more difficult to recognize persons who have changed clothes or hair styles in the surveillance videos, even by human eyeballing. To minimize mistakes and errors, before labeling, we need to pay vast efforts to watch thousands of hours of videos to familiarize the persons recorded in the videos.
Further, it is extremely challenging when persons wear masks, hats, or heavy winter clothes.
During labeling, we inspected the labeled persons frequently to avoid duplicate identities assigned to the same person.
{As a consequence, the quality of our annotations is satisfactory.}

\vspace{0.1cm}
\noindent{\bf Pedestrian detection }
After labeling, 
we used 
Faster RCNN~\cite{ren2015faster} to detect bounding boxes. For each selected bounding box, we annotated person ID, camera ID, tracklet ID, and {time stamp}. {Finally, we detected and 
blurred the face area for privacy protection.}

\begin{figure*}[!th]
	\centering	
	\includegraphics[width=\linewidth]{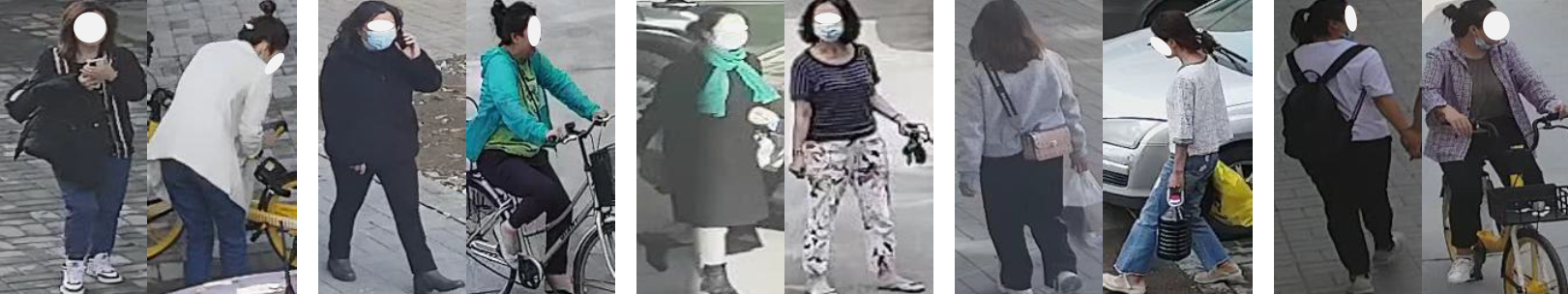}
	\caption{Sample pairs of the specific persons with simultaneous clothes and hair style changes in the \name{} dataset. For each identity, only two cases are selected randomly. Best viewed in color.}
	\label{fig:hair-style}
\end{figure*}

\begin{table*}[!t]
  \caption{Comparison with existing long-term image-based re-id datasets with clothes change.
  `Fas.': Faster RCNN~\cite{ren2015faster}, `Mas.': Mask RCNN~\cite{he2017mask}, `ind.': indoor, `out.': outdoor, `CD': {\bf cross-day}, `CM': {\bf cross-month}, `CS': {\bf cross-season}, `CY': {\bf cross-year}, `spr.': {\bf spring}, `sum.': {\bf summer}, `aut.': {\bf autumn}, `win.': {\bf winter}, 'sim.': simulated, `sur.': surveillance, '-': unknown.}
  \label{comparison-with-cloth-change-dataset}
  \centering
  \resizebox{\linewidth}{!}{
  \begin{tabular}{l r r r l l l l l l l l | l l l l}
    \hline
\multirow{2}{*}{Dataset} & \multirow{2}{*}{\# Person} & \multirow{2}{*}{\# Box} & \multirow{2}{*}{\# Cam} & \multirow{2}{*}{Source} & \multirow{2}{*}{Detector} & \multirow{2}{*}{Scene} & \multicolumn{5}{c|}{Time Range} & \multicolumn{4}{c}{Cloth Style} \\
\cline{8-12} \cline{13-16}
 & & & & & & & course & CD & CM & CS & CY & spr. & sum. & aut. & win. \\

    \hline
    \hline
Real28~\cite{wan2020person} & 28 & 4.3K & 4 & sim. &  Mas. & out., ind. & 3 days & $\checkmark$ & &  & & & $\checkmark$ & & \\

NKUP~\cite{wang2020benchmark} & 107 & 9.7K & 15 & sim. & -  & out., ind. & 4 months & $\checkmark$ & $\checkmark$ & $\checkmark$ & & & & $\checkmark$ & $\checkmark$ \\

LTCC~\cite{qian2020long} & 152 & 17K & 12 & sim. & Mas.  & ind. & 2 months & $\checkmark$ & $\checkmark$ &  & & & $\checkmark$ & & \\
 
PRCC~\cite{yang2019person} & 221 & 33K & 3 & sim. &  - & ind. & - & $\checkmark$ & &  & & & $\checkmark$ & & \\
 
 
 
 \hline
 \hline
\name & {\bf 1,121} & {\bf 178K} & {\bf 17} & {\bf sur.} & Fas.  & out. & {\bf 12 months} & $\checkmark$ & $\checkmark$ & $\checkmark$ & $\checkmark$ & $\checkmark$ & $\checkmark$ & $\checkmark$ & $\checkmark$ \\
    \hline
  \end{tabular}
  }
\end{table*}

\begin{table*}[!t]
  \caption{Comparison with conventional short-term image-based Re-ID datasets without clothes change. (`Fas.': Faster RCNN~\cite{ren2015faster}, `DPM': Deformable Part Model~\cite{felzenszwalb2009object}, `ind.': indoor, `out.': outdoor)}
  \label{comparison-with-traditional-dataset}
  \centering
  \resizebox{\linewidth}{!}{
  \begin{tabular}{l l l l l l l l l l}
    \hline

Dataset & \name & MSMT17~\cite{wei2018person} & Duke~\cite{zheng2017unlabeled} & Market~\cite{zheng2015scalable} & CUHK03~\cite{li2014deepreid} & CUHK01~\cite{li2012human} & VIPeR~\cite{gray2008viewpoint} & PRID~\cite{hirzer2011person} & CAVIAR~\cite{cheng2011custom} \\

    \hline
    \hline
\# Person & 1,121 & 4,101 & 1,812 & 1,501 & 1,467 & 971 & 632 & 934 & 72 \\
\# Bbox & {\bf 178K} & 126K & 36K & 32K & 28K & 3.8K & 1.2K & 1.1K & 0.6K \\
\# Camera & {\bf 17} & 15 & 8 & 6 & 2 & 10 & 2 & 2 & 2 \\
Detector & Fas. & Fas. & hand & DPM & DPM, hand & hand & hand & hand & hand \\
Scene & out. & out. \& ind. & out. & out. & ind. & ind. & out. & out. & ind. \\
    \hline
  \end{tabular}
  }
\end{table*}





\vspace{0.1cm}
\noindent{\bf Data statistics }
All person identities were captured by at least two cameras with most seen by $2 \sim 6$ cameras (as shown in Figure~\ref{fig:pid-cid-bar}). Figure~\ref{fig:hour-bbox-bar} indicates that the labeled bounding boxes are distributed from $6~am$ to $9~pm$. 
As illustrated in Figure~\ref{fig:season-bbox-pie}, the bounding box ratios of persons wearing spring\&autumn, summer, and winter clothes are $59.97\%$, $32.99\%$, and $7.03\%$, respectively.
More detailed statistics can be found in Figure~\ref{fig:statistic-analysis}.

\vspace{0.1cm}
\noindent{\bf Data splitting}
We shuffled all our collected identities, and then orderly picked $450$, $150$, and $521$ IDs for training, validation, and test, respectively.
In validation and test sets, {given a tracklet}, we randomly chose $\sim5$ bounding boxes as queries/probes, and the remaining boxes were split into the gallery.
Details were summarized in Table~\ref{dataset-split-table}.

\vspace{0.1cm}
\noindent{\bf Diversity and challenge}
As aforementioned, this wide ($14$ hectares) and dense block 
provides various identities (as shown in Figure~\ref{fig:identity-diversity}), and middle temperate continental monsoon climate causes diverse clothes changes (as demonstrated in Figure~\ref{fig:identity-diversity}).
Our long-term video collection makes full use of these characteristics of this venue and the climate.
We observed that obvious hair-style changes often happened in long-term surveillance videos. In Figure~\ref{fig:hair-style}, we present some random cases with simultaneous clothes and hair style changes.
It is interesting to see that hair style changes should also be considered in long-term \reid{}, as this might lead to non-neglectable appearance alternation.~\footnote{More illustrations are provided in the appendix.}

\begin{figure*}[!t]
	\centering
	\begin{subfigure}{0.4\linewidth}
		\includegraphics[width=\linewidth]{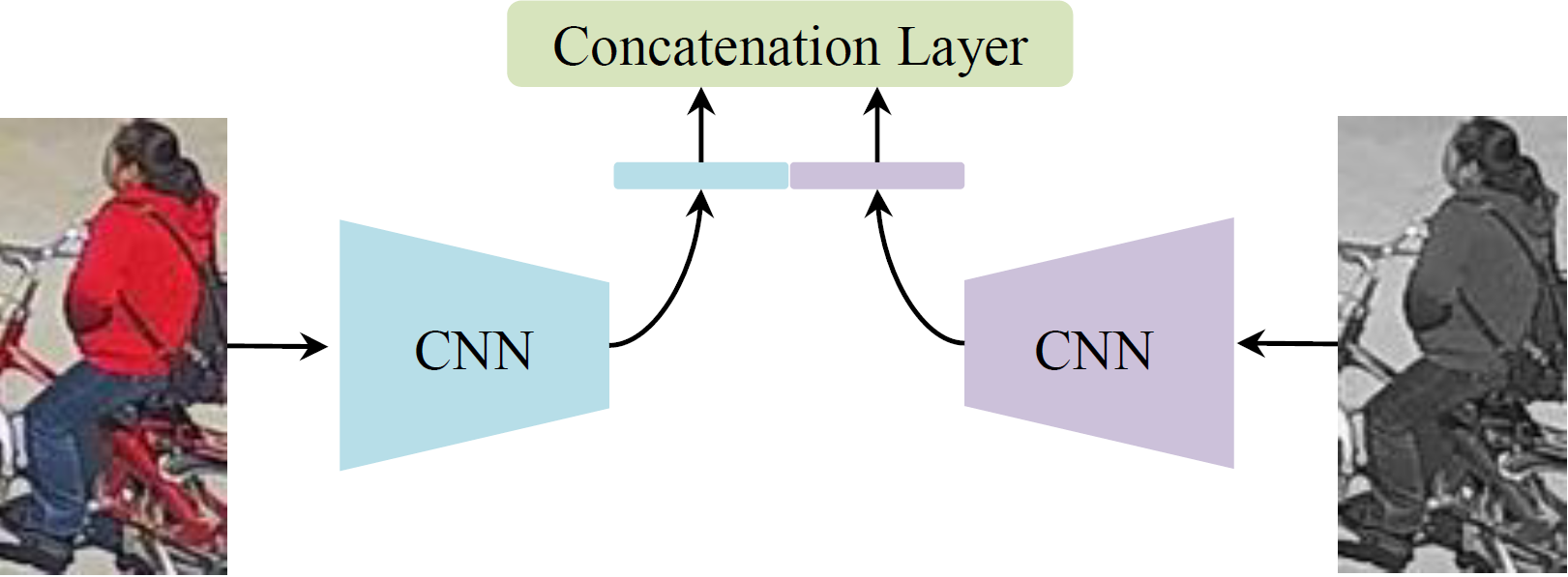}
		\caption{~}
		\label{fig:cnn-fusion}
	\end{subfigure}
	\hspace{2cm}
    \begin{subfigure}{0.45\linewidth}
		\includegraphics[width=\linewidth]{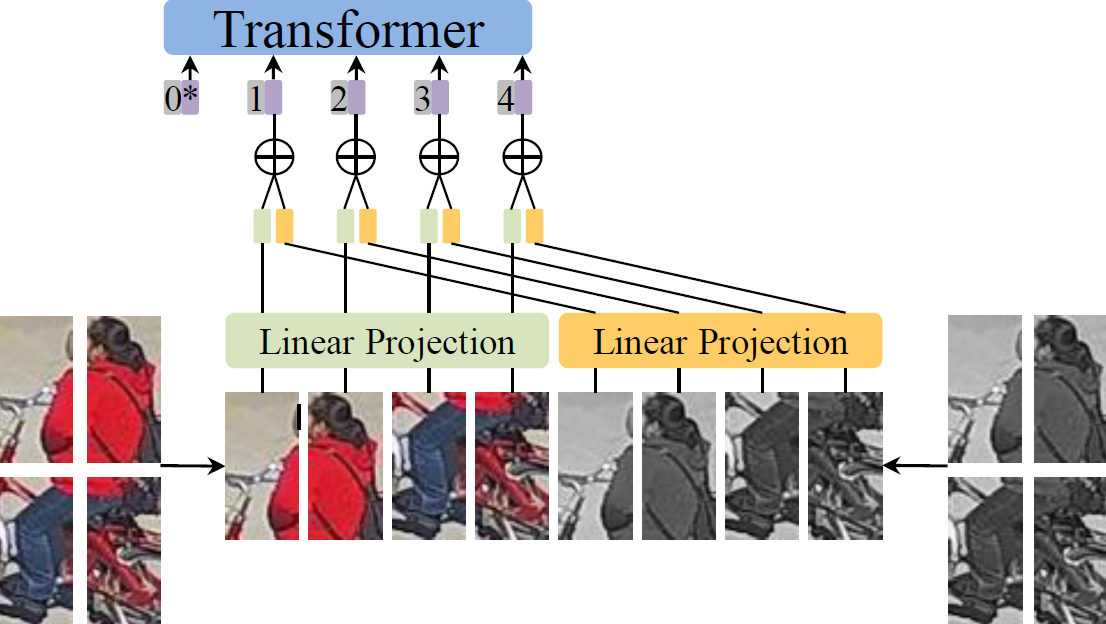}
		\caption{~}
		\label{fig:transformer-fusion}
	\end{subfigure}
		
	\caption{Two multimodal fusion strategies: (a) CNN-based late-fusion, (b) Transformer-based early-fusion.
	Both support more than two modalities.
	}
	\label{fig:fusion}
\end{figure*}

\subsection{Comparison with Existing Datasets}
We compare our \name{} with 
existing \reid{} datasets with and without clothes change.
We only discuss the publicly available and non-synthetic datasets.
As summarized in Table~\ref{comparison-with-cloth-change-dataset}, compared with existing long-term datasets, \name{} has the {\bf largest} number of identities, cameras, and bounding boxes, with the {\bf longest} time span. Besides, it is the only dataset offering four seasonal dressing styles. As seen in Table~\ref{comparison-with-traditional-dataset}, \name{} still has the {\bf largest} number of cameras and bounding boxes even in comparison to traditional short-term \reid{} datasets.

\section{Multimodal Fusion for Tackling the Clothes Change Challenge}
{
Conventional \reid{} methods mostly use the color information \cite{gong2014person,wei2018person,zheng2016person}.
At presence of clothes change, this may be insufficient 
given that different clothes could vary arbitrarily in color.
To tackle this challenge, we consider a simultaneous use of multiple modalities so that additional information irrelevant to the clothes' appearance can be leveraged concurrently.
In particular, we consider three more modalities,
grayscale images, edge maps, and skeleton key points,
for providing the potentially useful knowledge about body contour and part proportion.
For cost-effective multimodal fusion,
we explore two strategies:
(i) {\bf CNN-based late-fusion} (Figure \ref{fig:cnn-fusion}): Multiple input modalities are separately encoded by a multi-branch network, followed by feature concatenation.
(ii) {\bf Transformer-based early-fusion} (Figure \ref{fig:transformer-fusion}): The patch tokens from multiple modalities are fused by weighted sum in a position-wise manner, followed by self-attention representation learning.
}


\section{Experiments}
\label{sec:experiments}

We evaluated common deep CNN models and state-of-the-art re-id methods on \name{}. 

\vspace{0.1cm}
\noindent{\bf Protocols and metrics}
In the traditional short-term person \reid{}, it is assumed that the appearance of a specific person does not change across time and space. However, this assumption often does not hold for the long-term setting.
Hence, in our benchmark we allow the true matches 
coming from the same camera as the probe/query image but from a different time and trajectory.
%
Following~\cite{zheng2015scalable,subramaniam2016deep,wei2018person,ge2018fd, eom2019learning,ge2020self}, we used both Cumulated Matching
Characteristics (CMC) and mean average precision (mAP) as accuracy metrics.


%
\vspace{0.1cm}
\noindent{\bf Implementation details }
For a fair comparison, all experiments were implemented in the same software and hardware platform. 
In training only random horizontal flipping was used for data augmentation. For reproducibility, we conducted a unified early stop strategy for all experiments. We empirically adopted mAP on evaluation set as early stop metric and set patience as $30$ epochs, thus the checkpoints with the highest validation
performance were chosen to report test performance.
As all the backbones were initialized by the weights pretrained on ImageNet~\cite{imagenet_cvpr09}, we empirically used an initial
learning rate of $1e^{-4}$, multiplied by a decay factor $0.1$ every
$20$ epochs.
ReIDCaps~\cite{huang2019beyond} and BNNeck \reid{}~\cite{luo2020strong} use the customized loss functions, while the other models were trained by minimizing the softmax based cross-entropy loss. During evaluating and testing, we extracted the features of bounding boxes from the penultimate layer to conduct \reid{} matching. Edge Box~\cite{ZitnickECCV14edgeBoxes} and Open Pose~\cite{cao2019openpose} toolboxes were used to extract edge maps and detect body keypoints ($15$ keypoints, $54D$ vector), respectively. For the multimodal methods, body keypoint vectors were passed into a module of two fully-connected layers with Batch Normalization~\cite{ioffe2015batch} and ReLU~\cite{glorot2011deep} activations, while CNNs were used to encode RGB images, grayscale images, and edge maps.

\begin{table*}[!t]
\tiny
  \caption{{Unimodal results} on the test set of \name. Both rank accuracy ($\%$) and mAP ($\%$) are reported. (`R': RGB, `G': Grayscale, `E': Edge map, `K': Body key point, `Mod.': Modalities)}
  \label{table:baselines}
  \centering
  \resizebox{\linewidth}{!}{
  \begin{tabular}{l l l c c c c c c}
    \hline

\multirow{2}{*}{Model} & \multicolumn{2}{c}{Input} &  \multirow{2}{*}{Batch size} & \multicolumn{4}{c}{Rank} & \multirow{2}{*}{mAP} \\
\cline{2-3} \cline{5-8}
 & Mod. & Resolution &   & @1 & @5 & @10 & @20 &  \\
    \hline
    \hline
\#1 ResNet18~\cite{he2016deep} & R & 256$\times$128  &  256 & 34.45    &  46.01 & 51.72  & 58.26  &  08.44 \\

\#2 ResNet18~\cite{he2016deep} & G & 256$\times$128  &  256 &  26.61 & 39.02  & 45.45  &  53.06 &  05.49 \\


\#3 ResNet34~\cite{he2016deep} & R & 256$\times$128 &  256  & 35.21 &  47.37 &  53.61 &  60.03 & 09.49 \\

\#4 ResNet34~\cite{he2016deep} & G & 256$\times$128  &  256  & 28.60 & 41.53  &  47.98 & 54.87  & 06.39 \\


\#5 ResNet50~\cite{he2016deep} & R & 256$\times$128  &  192 & 36.62 & 49.88  &  55.46 & 61.92  & 09.62 \\

\#6 ResNet50~\cite{he2016deep} & G & 256$\times$128  &  192 & 30.04 & 43.12  & 49.82  & 57.04  & 06.96 \\

\#7 ResNet50~\cite{he2016deep} & E & 256$\times$128  &  192 & 16.05 &  28.51 &  35.59  &  43.28 & 03.17 \\


\#8 ResNext50\_32x4d~\cite{he2016deep} & R & 256$\times$128 & {128} & {36.94} & {48.87}  & {54.55}  & {60.50}  & {10.09} \\
\#9 Wide\_ResNet50\_2~\cite{he2016deep} & R & 256$\times$128 & {128} & {33.68} & {45.57}  & {51.45}  & {57.72}  & {08.71} \\

\#10 ResNet101~\cite{he2016deep} & R & 256$\times$128  &  128 & 39.31 & 51.65  & 57.36  & 63.72  & 11.00 \\

\#11 ResNext101\_32x8d~\cite{he2016deep} & R & 256$\times$128 & {64} & {40.93} & {52.68}  & {58.05}  & {64.21}  & {11.07} \\
\#12 Wide\_ResNet101\_2~\cite{he2016deep} & R & 256$\times$128 & {64} & {34.33} & {46.46}  & {52.42}  & {58.69}  & {09.03} \\
\#13 ResNet152~\cite{he2016deep} & R & 256$\times$128  &  96 & 39.84 & 52.51  & 58.35  &  64.75 & 11.49 \\

\#14 MobileNetv2~\cite{sandler2018mobilenetv2} & R & 256$\times$128  &  256 & 33.71 & 46.51  & 52.72  & 59.45  & 07.95 \\

\#15 MobileNetv3 Small~\cite{howard2019searching}     & R & 256$\times$128 & {768} & {28.73} & {40.87}  & {47.24}  & {54.12}  & {06.09} \\
\#16 MobileNetv3 Large~\cite{howard2019searching}     & R & 256$\times$128 & {320} & {33.08} & {44.99}  & {51.32}  & {58.14}  & {07.56} \\
\#17 GoogLeNet~\cite{szegedy2015going}      & R & 299$\times$299 & {96} & {29.02} & {40.83}  & {46.75}  & {53.65}  & {07.24} \\
\#18 Inceptionv3~\cite{Szegedy_2016_CVPR}  & R & 299$\times$299  &  96 &  35.02 &  47.71 &  53.91 &  60.64 & 08.85 \\


{\#19 DenseNet121}~\cite{huang2017densely} & R & 256$\times$128  &  128 & 38.26 & 50.27  &  55.91 &  62.40 & 09.12 \\
{\#20 DenseNet121}~\cite{huang2017densely} & {G} & 256$\times$128  &  128 & {30.64} & {43.02}  & {49.41}  & {56.44}  & {06.14}  \\
{\#21 DenseNet121}~\cite{huang2017densely} & {E} & 256$\times$128  &  128 &  {18.21} & {30.40}  & {36.86}  & {44.04}  & {02.88} \\
{\#22 DenseNet121}~\cite{huang2017densely} & R & {224$\times$224}  &  {64} & {39.92}  &  {51.76}  &  {57.21}  &   {62.98} & {09.99}  \\
\#23 DenseNet161~\cite{huang2017densely} & R & 256$\times$128  &  64 & 45.92 &  56.72 & 61.79  & 67.41  & 12.30 \\
\#24 DenseNet169~\cite{huang2017densely} & R & 256$\times$128  &  96 & 43.40 &  54.80 &  60.11 &  65.90 & 11.25 \\

\#25 DenseNet201~\cite{huang2017densely} & R & 256$\times$128  &  64 & 44.98 & 56.13  & 61.32  & 66.98  & 11.71 \\
\#26 ShuffleNetv2\_x0.5~\cite{ma2018shufflenet}      & R & 256$\times$128 & {1024} & {17.21} & {27.69}  & {33.48}  & {40.53}  & {04.00} \\
\#27 ShuffleNetv2\_x1.0~\cite{ma2018shufflenet}      & R & 256$\times$128 & {768} & {20.22} & {32.95}  & {39.58}  & {46.89}  & {04.60} \\
\#28 SqueezeNet1\_0~\cite{iandola2016squeezenet}      & R & 256$\times$128 & {384} & {23.38} & {33.17}  & {38.64}  & {44.34}  & {04.78} \\
\#29 SqueezeNet1\_1~\cite{iandola2016squeezenet}      & R & 256$\times$128 & {640} & {23.52} & {33.54}  & {39.26}  & {45.59}  & {05.02} \\
\#30 MnasNet0\_5~\cite{tan2019mnasnet} & R & 256$\times$128 & 512 & {17.52} & {29.17}  & {36.20}  & {43.51}  & {02.79} \\
\#31 MnasNet1\_0~\cite{tan2019mnasnet} & R & 256$\times$128 & {256} & {30.63} & {44.02}  & {50.40}  & {57.56}  & {06.25} \\
\hline
\hline
\#32 SCNet50~\cite{liu2020improving}      & R & 256$\times$128 & {160} & {35.07} & {47.73}  & {54.32}  & {60.73}  & {09.53} \\
\#33 SCNet50\_v1d~\cite{liu2020improving}      & R & 256$\times$128 & {128} & {37.46} & {50.20}  & {56.13}  & {62.36}  & {09.42} \\
\#34 SCNet101~\cite{liu2020improving}      & R & 256$\times$128 & {96} & {37.25} & {50.09}  & {56.28}  & {63.16}  & {10.35} \\
\hline
\hline
\#35 OSNet ibn x1.0~\cite{zhou2021osnet} & R & 256$\times$128 & {96} & {42.75} & {54.91}  & {60.82}  & {66.80}  & {10.97} \\
\#36 OSNet x1.0~\cite{zhou2021osnet}     & R & 256$\times$128 & 96 & {39.65} & {52.22}  & {58.32}  & {64.23}  & {10.34} \\
\#37 OSNet x0.75~\cite{zhou2021osnet}     & R & 256$\times$128 & 160 & {39.96} & {51.89}  & {57.69}  & {63.85}  & {09.92} \\
\#38 OSNet x0.5~\cite{zhou2021osnet}     & R & 256$\times$128 & 256 & {38.09} & {51.08}  & {56.79}  & {63.27}  & {09.59} \\
\#39 OSNet x0.25~\cite{zhou2021osnet}     & R & 256$\times$128 & 512 & {34.94} & {47.70}  & {54.10}  & {60.77}  & {08.62} \\

\#40 BNNeck \reid{} ResNet18~\cite{luo2020strong}     & R & 256$\times$128 & {224} & {38.17} & {51.93}  & {58.08}  & {64.70}  & {09.51} \\
\#41 BNNeck \reid{} ResNet34~\cite{luo2020strong}     & R & 256$\times$128 & {128} & {40.06} & {53.49}  & {59.55}  & {66.25}  & {10.52} \\
\#42 BNNeck \reid{} ResNet50~\cite{luo2020strong}      & R & 256$\times$128 & {56} & {47.45} & {59.47}  & {65.19}  & {71.10}  & {12.98} \\
\#43 BNNeck \reid{} ResNet50~\cite{luo2020strong}      & G & 256$\times$128 & {56} & {40.02} & {54.04}  & {60.19}  & {67.09}  & {09.43} \\
\#44 BNNeck \reid{} ResNet50~\cite{luo2020strong}      & E & 256$\times$128 & {56} & {21.75} & {35.99}  & {43.11}  & {50.81}  & {03.67} \\
\#45 BNNeck \reid{} ResNet101~\cite{luo2020strong}     & R & 256$\times$128 & {40} & {48.10} & {60.70}  & {66.10}  & {72.06}  & {13.72} \\
\#46 BNNeck \reid{} ResNet152~\cite{luo2020strong}     & R & 256$\times$128 & 28 & {50.29} & {62.27}  & {67.85}  & {73.63}  & {14.59} \\
\#47 BNNeck \reid{} DenseNet121~\cite{luo2020strong}     & R & 256$\times$128 & 40 & {47.86} & {60.47}  & {65.88}  & {71.64}  & {13.41} \\
\hline
\hline
\#48 ReIDCaps~\cite{huang2019beyond} (DenseNet121) & R & 224$\times$224  &  24 & 44.29 & 56.44  & 62.01  & 68.01  & 13.25 \\
\#49 ReIDCaps~\cite{huang2019beyond} (ResNet50) & R & {224$\times$224}  &  {32} & {39.49} & {52.28}  &  {58.68} & {64.99}  & {11.33} \\
   
{\#50 ReIDCaps~\cite{huang2019beyond} (no auxiliary)} & R & 224$\times$224  &  24 & {35.41} & {46.66}  & {52.09}  & {58.13}  & {09.25} \\
 
{\#51 ReIDCaps~\cite{huang2019beyond} (no capsule)} & R & {224$\times$224}  &  80 & {39.38} & {51.86}  & {57.82}  & {64.44}  & {11.16} \\
\hline
\hline

\#52 ViT B16~\cite{dosovitskiy2020image}            & R & {256$\times$128} & 64 & {49.78} & {61.81}  & {67.38}  & {72.92}  & {\bf 14.98} \\ 

\#53 ViT B16~\cite{dosovitskiy2020image}                  & G\cut{ray} & {256$\times$128} & 64 & 38.52 & 51.85 & 58.32 & 65.12 & 10.63 \\

\#54 DeiT~\cite{touvron2021training}     & R & 256$\times$128 & 64 & {44.43} & {56.25} & {61.82} & {67.46} & {13.72} \\
    \hline
  \end{tabular}
  }
\end{table*}

\begin{table*}[!t]
\tiny
  \caption{{Multimodal results} on the test set of \name. Both rank accuracy ($\%$) and mAP ($\%$) are reported. (`R': RGB, `G': Grayscale, `E': Edge map, `K': Body key point, `2br': Two branches, `3br': Three branches, `Mod.': Modalities, `Dim.': Dimensions)}
  \label{table:multi-modal-baselines}
  \centering
  \resizebox{\linewidth}{!}{
  \begin{tabular}{l l l c c c c c c}
    \hline

\multirow{2}{*}{Model} & \multicolumn{2}{c}{Input} &  \multirow{2}{*}{Batch size} & \multicolumn{4}{c}{Rank} & \multirow{2}{*}{mAP} \\
\cline{2-3} \cline{5-8}
 & Mod. & Dim. &   & @1 & @5 & @10 & @20 &  \\
    \hline
    \hline

\#55 2br ResNet50 & R, K & 256$\times$128  &  192 & 36.53 &  48.87 &  54.86 &  61.47 & 09.54 \\

\#56 2br ResNet50 & R, E & 256$\times$128  &  96 & 40.26 & 52.91  & 59.11  & 65.47  & 10.43 \\

\#57 2br ResNet50 & R, G & 256$\times$128  &  96 & 40.52 &  53.65  &  59.61 &  65.60 & 10.22 \\

\#58 3br ResNet50 & R, G, E & 256$\times$128  &  64 & 41.67 & 54.28  & 60.04   &  66.37 & 11.03 \\

\hline
\hline

\#59 2br DenseNet121 & R, E & 256$\times$128  &  64 & 44.55 & 56.40  &  62.03 &  67.85 & 11.21 \\

\#60 2br DenseNet121 & R, G & 256$\times$128  &  64 & 44.80 &  56.79 &  62.48 & 68.06  & 11.36 \\

\#61 3br DenseNet121 & R, G, E & 256$\times$128  &  32 & 45.36 & 57.36  & 62.91  & 69.29  & 11.73 \\
    \hline
    \hline
\#62 2br BNNeck \reid{} ResNet50~\cite{luo2020strong}      & R, G & 256$\times$128 & {28} & {46.62} & {59.72}  & {65.58}  & {71.48}  & {13.12} \\

\hline
\hline


\#63 \cut{2br} ViT B16~\cite{dosovitskiy2020image}    & 1* R + 0.1 * G      &  256$\times$128  &  32 & {47.83} &59.29&64.59&70.23&15.13\\
\#64 \cut{2br} ViT B16~\cite{dosovitskiy2020image}    & 1* R + 0.3 * G      &  256$\times$128  &  32& {48.00} &59.47&64.65&70.04& {\bf 15.19} \\

    \hline
  \end{tabular}
  }
\end{table*}

\begin{figure*}[!t]
	\centering
	\begin{subfigure}{0.7\linewidth}
		\includegraphics[width=\linewidth]{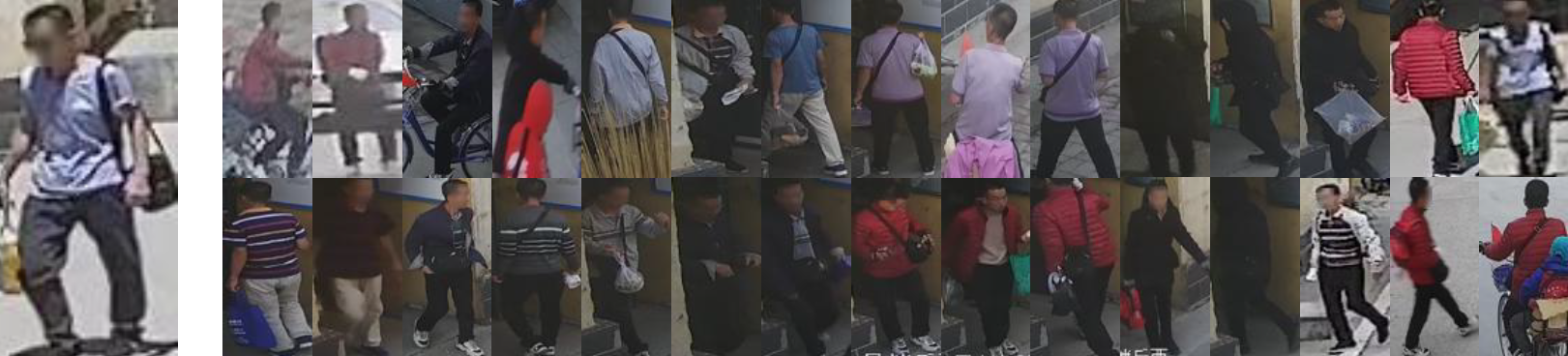}
		\caption{~}
		\label{fig:long-tail-1}
	\end{subfigure}
    \begin{subfigure}{0.25\linewidth}
		\includegraphics[width=\linewidth]{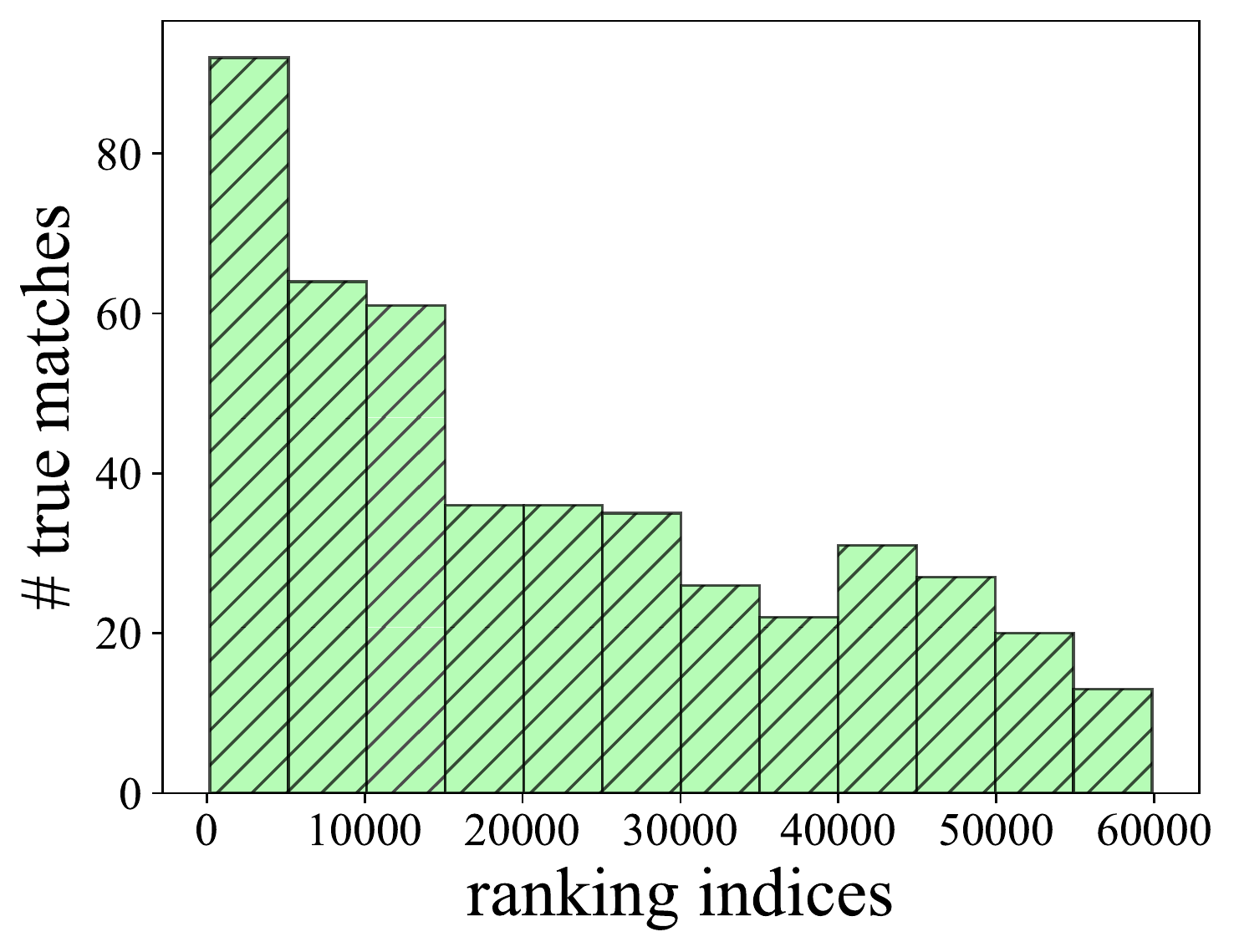}
		\caption{~}
		\label{fig:long-tail-2}
	\end{subfigure}
		
	\caption{A qualitative analysis of person \reid{} with clothes change.
	(a) {\em Left}: A random query image; {\em Right}: $30$ randomly selected true matches from the gallery. (b) The rank histogram of all the true matches.
	}
	\label{fig:long-tail}
\end{figure*}

\vspace{0.1cm}
\noindent{\bf Result analysis }
{We report the results of unimodal and multimodal methods 
in Table~\ref{table:baselines} and Table~\ref{table:multi-modal-baselines}, respectively.
We consider five groups of methods:
(1) \#1 to \#31: Common CNNs used in \reid{}; 
(2) \#32 to \#34: Latest CNNs;
(3) \#35 to \#47: Several state-of-the-art short-term \reid{} methods;
(4) \#48 to \#51: A state-of-the-art long-term \reid{} method ReIDCaps~\cite{huang2019beyond};
(5) \#52 to \#54: Vision Transformer (ViT)~\cite{dosovitskiy2020image} and its variant DeiT~\cite{touvron2021training}.
We draw the following main observations:}\\
{
{\bf(i)} For the results with RGB input, deeper models are usually superior than shallower ones as expected. For example, we see a clear upward trend from \#1 ResNet18 (rank@1: $34.45\%$, mAP: $08.44\%$) to \#13 ResNet152 (rank@1: $39.84\%$, mAP: $11.49\%$).
It is observed that ResNets and DenseNets outperform MobileNetv2 and Inceptionv3 by a clear margin.
Lightweight networks tend to underperform, \eg, ShuffleNetv2.
In general, these observations are largely consistent with conventional
short-term \reid{} results. \\
{\bf(ii)} As \reid{} specific models, BNNeck \reid{} and ReIDCaps achieve good performances. Similarly, both benefit from deeper networks, as seen from \#40 to \#46, \#49 to \#48. \\
{\bf(iii)} ViT obtains the best mAP score among all the unimodal methods (\#52), indicating that self-attention representation is superior in dealing with clothes change challenge over CNN models.
\cut{However, the benefit of using capsule layers is considerable
(\#12 vs. \#16) \todo.}
\\
{\bf(iv)} With different modalities (\#5 RGB, \#6 grayscale, and \#7 edge map) for ResNet50,
it is observed that RGB is the best.
Plausible reasons for lower performance with grayimages include:
Although being more tolerant with clothes change,
they offer less information than colorful ones.
Besides, CNNs are pretrained on colorful ImageNet images.
Note, we did not test keypoint modality in isolation,
as it fails on $13.45\%$ ($23K$ out of $178K$) of person images.
\\
%
{\bf(v)} From the multimodal results, we see that both edge maps and grayscale images are useful for performance by feature late-fusion (\#5 vs. \#56, \#57, and \#58).
This validates our hypothesis that jointly using multiple modalities
with specific appearance indeed
provides extra information for tackling the clothes change challenge.
It is expected that higher-quality edge maps would contribute more to the performance.
Again, we find that using stronger networks (\eg, \#59, \#60, \#61) can further improve the results in the multimodal case.
On the other hand, multimodal fusion also helps state-of-the-art
\reid{} models (\eg, \#42 vs. \#62).
This indicates good complementary effect between multimodal fusion and model refinement. 
With the proposed fusion method on RGB and grayscale images, 
interestingly the standard ViT model achieves the best mAP scores, surpassing a wide variety of CNNs and CNN based \reid{} methods by a clear margin.
This indicates that Transformers might be a stronger 
architecture for long-term person \reid{}, with promising investigation
potentials for the future.
}\\
{{\bf(vi)} Unlike short-term person \reid{}, we see that in long-term setting mAP scores are significantly lower than rank@1 rates, a consistent finding as in \cite{huang2019beyond, yang2019person}.
This is the most fundamental challenge with clothes change,
as there is little correlation between different clothes a specific person would wear over space and time. 
It is also not an easy task for human beings.
%
Due to high clothes diversity in our \name{} dataset, this challenge becomes more severe. For example,  
given a random query image (Figure~\ref{fig:long-tail-1}), 
the true matches with varying clothes
often form a long tail distribution in rank (Figure~\ref{fig:long-tail-2}).
This contributes to a low mAP score.
}

\section{Conclusions}
\label{sec:conclusions}
In this paper, we have introduced the only realistic, large-scale long-term person \reid{} datasets with natural clothes change,
named \name{}.
This aims for facilitating the research towards more realistic person \reid{} applications over space and time.
Compared with existing alternative datasets, 
it is established uniquely using real-world video sources without any artificial simulation.
Constructed with a huge amount of annotation efforts, \name{} contains the largest number of cameras, identities, and bounding boxes, and covers the longest time period and native appearance change. 
We conduct extensive experiments
using a wide variety of CNN and Transformer models and state-of-the-art \reid{} methods,
offering a set of rich baselines for future research works. 
{Further, we investigate multimodal fusion strategies 
for overcoming the clothes change challenge, and achieve new 
state-of-the-art results on our \name{} dataset.}



\section{Future work}
\label{sec:future_work}
In the future efforts, we would try to extend this work mainly in following aspects:
(i) continually collecting video to provide longer term test bed, \eg, three years,
(ii) annotating more person identities to enlarge training/validation/testing subsets and provide more complex appearance changes,
(iii) accommodating more cameras to provide more view variety,
(iv) creating a video-based long-term \reid{} dataset.

{\small
\bibliographystyle{ieee_fullname}
\bibliography{egbib}

\begin{thebibliography}{10}\itemsep=-1pt

\bibitem{cao2019openpose}
Zhe Cao, Gines Hidalgo, Tomas Simon, Shih-En Wei, and Yaser Sheikh.
\newblock Openpose: realtime multi-person 2d pose estimation using part
  affinity fields.
\newblock {\em TPAMI}, 2019.

\bibitem{chang2018multi}
Xiaobin Chang, Timothy~M Hospedales, and Tao Xiang.
\newblock Multi-level factorisation net for person re-identification.
\newblock In {\em CVPR}, 2018.

\bibitem{chenyanbei2017person}
Yanbei Chen, Xiatian Zhu, and Shaogang Gong.
\newblock Person re-identification by deep learning multi-scale
  representations.
\newblock In {\em ICCVW}, 2017.

\bibitem{chen2017person}
Ying-Cong Chen, Xiatian Zhu, Wei-Shi Zheng, and Jian-Huang Lai.
\newblock Person re-identification by camera correlation aware feature
  augmentation.
\newblock {\em TPAMI}, 2017.

\bibitem{cheng2011custom}
Dong~Seon Cheng, Marco Cristani, Michele Stoppa, Loris Bazzani, and Vittorio
  Murino.
\newblock Custom pictorial structures for re-identification.
\newblock In {\em BMVC}, 2011.

\bibitem{cheng2020inter}
Zhiyi Cheng, Qi Dong, Shaogang Gong, and Xiatian Zhu.
\newblock Inter-task association critic for cross-resolution person
  re-identification.
\newblock In {\em CVPR}, 2020.

\bibitem{imagenet_cvpr09}
J. Deng, W. Dong, R. Socher, L.-J. Li, K. Li, and L. Fei-Fei.
\newblock {ImageNet: A Large-Scale Hierarchical Image Database}.
\newblock In {\em CVPR}, 2009.

\bibitem{dosovitskiy2020image}
Alexey Dosovitskiy, Lucas Beyer, Alexander Kolesnikov, Dirk Weissenborn,
  Xiaohua Zhai, Thomas Unterthiner, Mostafa Dehghani, Matthias Minderer, Georg
  Heigold, Sylvain Gelly, et~al.
\newblock An image is worth 16x16 words: Transformers for image recognition at
  scale.
\newblock {\em arXiv preprint arXiv:2010.11929}, 2020.

\bibitem{eom2019learning}
Chanho Eom and Bumsub Ham.
\newblock Learning disentangled representation for robust person
  re-identification.
\newblock In {\em NeurIPS}, 2019.

\bibitem{felzenszwalb2009object}
Pedro~F Felzenszwalb, Ross~B Girshick, David McAllester, and Deva Ramanan.
\newblock Object detection with discriminatively trained part-based models.
\newblock {\em TPAMI}, 2009.

\bibitem{ge2020mutual}
Yixiao Ge, Dapeng Chen, and Hongsheng Li.
\newblock Mutual mean-teaching: Pseudo label refinery for unsupervised domain
  adaptation on person re-identification.
\newblock {\em arXiv preprint arXiv:2001.01526}, 2020.

\bibitem{ge2018fd}
Yixiao Ge, Zhuowan Li, Haiyu Zhao, Guojun Yin, Shuai Yi, Xiaogang Wang, and
  Hongsheng Li.
\newblock Fd-gan: Pose-guided feature distilling gan for robust person
  re-identification.
\newblock In {\em NeurIPS}, 2018.

\bibitem{ge2020self}
Yixiao Ge, Feng Zhu, Dapeng Chen, Rui Zhao, and Hongsheng Li.
\newblock Self-paced contrastive learning with hybrid memory for domain
  adaptive object re-id.
\newblock In {\em NeurIPS}, 2020.

\bibitem{glorot2011deep}
Xavier Glorot, Antoine Bordes, and Yoshua Bengio.
\newblock Deep sparse rectifier neural networks.
\newblock In {\em AISTATS}, 2011.

\bibitem{gong2014person}
Shaogang Gong, Marco Cristani, Shuicheng Yan, and Chen~Change Loy.
\newblock {\em Person re-identification}.
\newblock Springer, January 2014.

\bibitem{gray2008viewpoint}
Douglas Gray and Hai Tao.
\newblock Viewpoint invariant pedestrian recognition with an ensemble of
  localized features.
\newblock In {\em ECCV}, 2008.

\bibitem{he2017mask}
Kaiming He, Georgia Gkioxari, Piotr Doll{\'a}r, and Ross Girshick.
\newblock Mask r-cnn.
\newblock In {\em ICCV}, 2017.

\bibitem{he2016deep}
Kaiming He, Xiangyu Zhang, Shaoqing Ren, and Jian Sun.
\newblock Deep residual learning for image recognition.
\newblock In {\em CVPR}, 2016.

\bibitem{hermans2017defense}
Alexander Hermans, Lucas Beyer, and Bastian Leibe.
\newblock In defense of the triplet loss for person re-identification.
\newblock {\em arXiv preprint arXiv:1703.07737}, 2017.

\bibitem{hirzer2011person}
Martin Hirzer, Csaba Beleznai, Peter~M Roth, and Horst Bischof.
\newblock Person re-identification by descriptive and discriminative
  classification.
\newblock In {\em SCIA}, 2011.

\bibitem{howard2019searching}
Andrew Howard, Mark Sandler, Grace Chu, Liang-Chieh Chen, Bo Chen, Mingxing
  Tan, Weijun Wang, Yukun Zhu, Ruoming Pang, Vijay Vasudevan, et~al.
\newblock Searching for mobilenetv3.
\newblock In {\em ICCV}, 2019.

\bibitem{huang2017densely}
Gao Huang, Zhuang Liu, Laurens Van Der~Maaten, and Kilian~Q Weinberger.
\newblock Densely connected convolutional networks.
\newblock In {\em CVPR}, 2017.

\bibitem{huang2019celebrities}
Yan Huang, Qiang Wu, Jingsong Xu, and Yi Zhong.
\newblock Celebrities-reid: A benchmark for clothes variation in long-term
  person re-identification.
\newblock In {\em IJCNN}, 2019.

\bibitem{huang2019beyond}
Yan Huang, Jingsong Xu, Qiang Wu, Yi Zhong, Peng Zhang, and Zhaoxiang Zhang.
\newblock Beyond scalar neuron: Adopting vector-neuron capsules for long-term
  person re-identification.
\newblock {\em TCSVT}, 2019.

\bibitem{iandola2016squeezenet}
Forrest~N Iandola, Song Han, Matthew~W Moskewicz, Khalid Ashraf, William~J
  Dally, and Kurt Keutzer.
\newblock Squeezenet: Alexnet-level accuracy with 50x fewer parameters and< 0.5
  mb model size.
\newblock {\em arXiv preprint arXiv:1602.07360}, 2016.

\bibitem{ioffe2015batch}
Sergey Ioffe and Christian Szegedy.
\newblock Batch normalization: Accelerating deep network training by reducing
  internal covariate shift.
\newblock In {\em ICML}, 2015.

\bibitem{jiao2018deep}
Jiening Jiao, Wei-Shi Zheng, Ancong Wu, Xiatian Zhu, and Shaogang Gong.
\newblock Deep low-resolution person re-identification.
\newblock In {\em AAAI}, 2018.

\bibitem{lan2017deep}
Xu Lan, Hanxiao Wang, Shaogang Gong, and Xiatian Zhu.
\newblock Deep reinforcement learning attention selection for person
  re-identification.
\newblock In {\em BMVC}, 2017.

\bibitem{li2018unsupervised}
Minxian Li, Xiatian Zhu, and Shaogang Gong.
\newblock Unsupervised person re-identification by deep learning tracklet
  association.
\newblock In {\em ECCV}, 2018.

\bibitem{li2019unsupervised}
Minxian Li, Xiatian Zhu, and Shaogang Gong.
\newblock Unsupervised tracklet person re-identification.
\newblock {\em TPAMI}, 2019.

\bibitem{li2012human}
Wei Li, Rui Zhao, and Xiaogang Wang.
\newblock Human reidentification with transferred metric learning.
\newblock In {\em ACCV}, 2012.

\bibitem{li2014deepreid}
Wei Li, Rui Zhao, Tong Xiao, and Xiaogang Wang.
\newblock Deepreid: Deep filter pairing neural network for person
  re-identification.
\newblock In {\em CVPR}, 2014.

\bibitem{li2018harmonious}
Wei Li, Xiatian Zhu, and Shaogang Gong.
\newblock Harmonious attention network for person re-identification.
\newblock In {\em CVPR}, 2018.

\bibitem{li2020learning}
Yu-Jhe Li, Zhengyi Luo, Xinshuo Weng, and Kris~M Kitani.
\newblock Learning shape representations for clothing variations in person
  re-identification.
\newblock {\em arXiv preprint arXiv:2003.07340}, 2020.

\bibitem{liu2020improving}
Jiang-Jiang Liu, Qibin Hou, Ming-Ming Cheng, Changhu Wang, and Jiashi Feng.
\newblock Improving convolutional networks with self-calibrated convolutions.
\newblock In {\em CVPR}, 2020.

\bibitem{luo2020strong}
Hao Luo, Wei Jiang, Youzhi Gu, Fuxu Liu, Xingyu Liao, Shenqi Lai, and Jianyang
  Gu.
\newblock A strong baseline and batch normalization neck for deep person
  re-identification.
\newblock {\em TMM}, 2020.

\bibitem{ma2018shufflenet}
Ningning Ma, Xiangyu Zhang, Hai-Tao Zheng, and Jian Sun.
\newblock Shufflenet v2: Practical guidelines for efficient cnn architecture
  design.
\newblock In {\em ECCV}, 2018.

\bibitem{prosser2010person}
Bryan~James Prosser, Wei-Shi Zheng, Shaogang Gong, Tao Xiang, Q Mary, et~al.
\newblock Person re-identification by support vector ranking.
\newblock In {\em BMVC}, 2010.

\bibitem{qian2020long}
Xuelin Qian, Wenxuan Wang, Li Zhang, Fangrui Zhu, Yanwei Fu, Tao Xiang, Yu-Gang
  Jiang, and Xiangyang Xue.
\newblock Long-term cloth-changing person re-identification.
\newblock In {\em ACCV}, 2020.

\bibitem{ren2015faster}
Shaoqing Ren, Kaiming He, Ross Girshick, and Jian Sun.
\newblock Faster r-cnn: Towards real-time object detection with region proposal
  networks.
\newblock In {\em NeurIPS}, 2015.

\bibitem{sandler2018mobilenetv2}
Mark Sandler, Andrew Howard, Menglong Zhu, Andrey Zhmoginov, and Liang-Chieh
  Chen.
\newblock Mobilenetv2: Inverted residuals and linear bottlenecks.
\newblock In {\em CVPR}, 2018.

\bibitem{subramaniam2016deep}
Arulkumar Subramaniam, Moitreya Chatterjee, and Anurag Mittal.
\newblock Deep neural networks with inexact matching for person
  re-identification.
\newblock In {\em NeurIPS}, 2016.

\bibitem{sun2018beyond}
Yifan Sun, Liang Zheng, Yi Yang, Qi Tian, and Shengjin Wang.
\newblock Beyond part models: Person retrieval with refined part pooling (and a
  strong convolutional baseline).
\newblock In {\em ECCV}, 2018.

\bibitem{szegedy2015going}
Christian Szegedy, Wei Liu, Yangqing Jia, Pierre Sermanet, Scott Reed, Dragomir
  Anguelov, Dumitru Erhan, Vincent Vanhoucke, and Andrew Rabinovich.
\newblock Going deeper with convolutions.
\newblock In {\em CVPR}, 2015.

\bibitem{Szegedy_2016_CVPR}
Christian Szegedy, Vincent Vanhoucke, Sergey Ioffe, Jon Shlens, and Zbigniew
  Wojna.
\newblock Rethinking the inception architecture for computer vision.
\newblock In {\em CVPR}, 2016.

\bibitem{tan2019mnasnet}
Mingxing Tan, Bo Chen, Ruoming Pang, Vijay Vasudevan, Mark Sandler, Andrew
  Howard, and Quoc~V Le.
\newblock Mnasnet: Platform-aware neural architecture search for mobile.
\newblock In {\em CVPR}, 2019.

\bibitem{touvron2021training}
Hugo Touvron, Matthieu Cord, Matthijs Douze, Francisco Massa, Alexandre
  Sablayrolles, and Herv{\'e} J{\'e}gou.
\newblock Training data-efficient image transformers \& distillation through
  attention.
\newblock In {\em ICML}, 2021.

\bibitem{wan2020person}
Fangbin Wan, Yang Wu, Xuelin Qian, Yixiong Chen, and Yanwei Fu.
\newblock When person re-identification meets changing clothes.
\newblock In {\em CVPRW}, 2020.

\bibitem{wang2018transferable}
Jingya Wang, Xiatian Zhu, Shaogang Gong, and Wei Li.
\newblock Transferable joint attribute-identity deep learning for unsupervised
  person re-identification.
\newblock In {\em CVPR}, 2018.

\bibitem{wang2020benchmark}
Kai Wang, Zhi Ma, Shiyan Chen, Jinni Yang, Keke Zhou, and Tao Li.
\newblock A benchmark for clothes variation in person re-identification.
\newblock {\em IJIS}, 2020.

\bibitem{wang2014person}
Taiqing Wang, Shaogang Gong, Xiatian Zhu, and Shengjin Wang.
\newblock Person re-identification by video ranking.
\newblock In {\em ECCV}, 2014.

\bibitem{wei2018person}
Longhui Wei, Shiliang Zhang, Wen Gao, and Qi Tian.
\newblock Person transfer gan to bridge domain gap for person
  re-identification.
\newblock In {\em CVPR}, 2018.

\bibitem{xiao2016learning}
Tong Xiao, Hongsheng Li, Wanli Ouyang, and Xiaogang Wang.
\newblock Learning deep feature representations with domain guided dropout for
  person re-identification.
\newblock In {\em CVPR}, 2016.

\bibitem{yang2019person}
Qize Yang, Ancong Wu, and Wei-Shi Zheng.
\newblock Person re-identification by contour sketch under moderate clothing
  change.
\newblock {\em TPAMI}, 2019.

\bibitem{ye2021deep}
Mang Ye, Jianbing Shen, Gaojie Lin, Tao Xiang, Ling Shao, and Steven~CH Hoi.
\newblock Deep learning for person re-identification: A survey and outlook.
\newblock {\em TPAMI}, 2021.

\bibitem{yin2020fine}
Jiahang Yin, Ancong Wu, and Wei-Shi Zheng.
\newblock Fine-grained person re-identification.
\newblock {\em IJCV}, 2020.

\bibitem{yu2017cross}
Hong-Xing Yu, Ancong Wu, and Wei-Shi Zheng.
\newblock Cross-view asymmetric metric learning for unsupervised person
  re-identification.
\newblock In {\em ICCV}, 2017.

\bibitem{yu2019unsupervised}
Hong-Xing Yu, Wei-Shi Zheng, Ancong Wu, Xiaowei Guo, Shaogang Gong, and
  Jian-Huang Lai.
\newblock Unsupervised person re-identification by soft multilabel learning.
\newblock In {\em CVPR}, 2019.

\bibitem{zhang2016learning}
Li Zhang, Tao Xiang, and Shaogang Gong.
\newblock Learning a discriminative null space for person re-identification.
\newblock In {\em CVPR}, 2016.

\bibitem{zhao2013unsupervised}
Rui Zhao, Wanli Ouyang, and Xiaogang Wang.
\newblock Unsupervised salience learning for person re-identification.
\newblock In {\em CVPR}, 2013.

\bibitem{zheng2015scalable}
Liang Zheng, Liyue Shen, Lu Tian, Shengjin Wang, Jingdong Wang, and Qi Tian.
\newblock Scalable person re-identification: A benchmark.
\newblock In {\em ICCV}, 2015.

\bibitem{zheng2016person}
Liang Zheng, Yi Yang, and Alexander~G Hauptmann.
\newblock Person re-identification: Past, present and future.
\newblock {\em arXiv preprint arXiv:1610.02984}, 2016.

\bibitem{zheng2012reidentification}
Wei-Shi Zheng, Shaogang Gong, and Tao Xiang.
\newblock Reidentification by relative distance comparison.
\newblock {\em TPAMI}, 2012.

\bibitem{zheng2017unlabeled}
Zhedong Zheng, Liang Zheng, and Yi Yang.
\newblock Unlabeled samples generated by gan improve the person
  re-identification baseline in vitro.
\newblock In {\em ICCV}, 2017.

\bibitem{zhong2020learning}
Zhun Zhong, Liang Zheng, Zhiming Luo, Shaozi Li, and Yi Yang.
\newblock Learning to adapt invariance in memory for person re-identification.
\newblock {\em TPAMI}, 2020.

\bibitem{zhou2021osnet}
Kaiyang Zhou, Yongxin Yang, Andrea Cavallaro, and Tao Xiang.
\newblock Learning generalisable omni-scale representations for person
  re-identification.
\newblock {\em TPAMI}, 2021.

\bibitem{zhu2021intra}
Xiangping Zhu, Xiatian Zhu, Minxian Li, Pietro Morerio, Vittorio Murino, and
  Shaogang Gong.
\newblock Intra-camera supervised person re-identification.
\newblock {\em IJCV}, 2021.

\bibitem{ZitnickECCV14edgeBoxes}
C.~Lawrence Zitnick and Piotr Doll\'ar.
\newblock Edge boxes: Locating object proposals from edges.
\newblock In {\em ECCV}, 2014.

\end{thebibliography}
}

\clearpage

\appendix

\paragraph{Supplementary figures} We provide more example images to further demonstrate the diversity of this dataset in person's appearance.

\begin{figure*}[!h]
	\centering	
	\includegraphics[width=\linewidth]{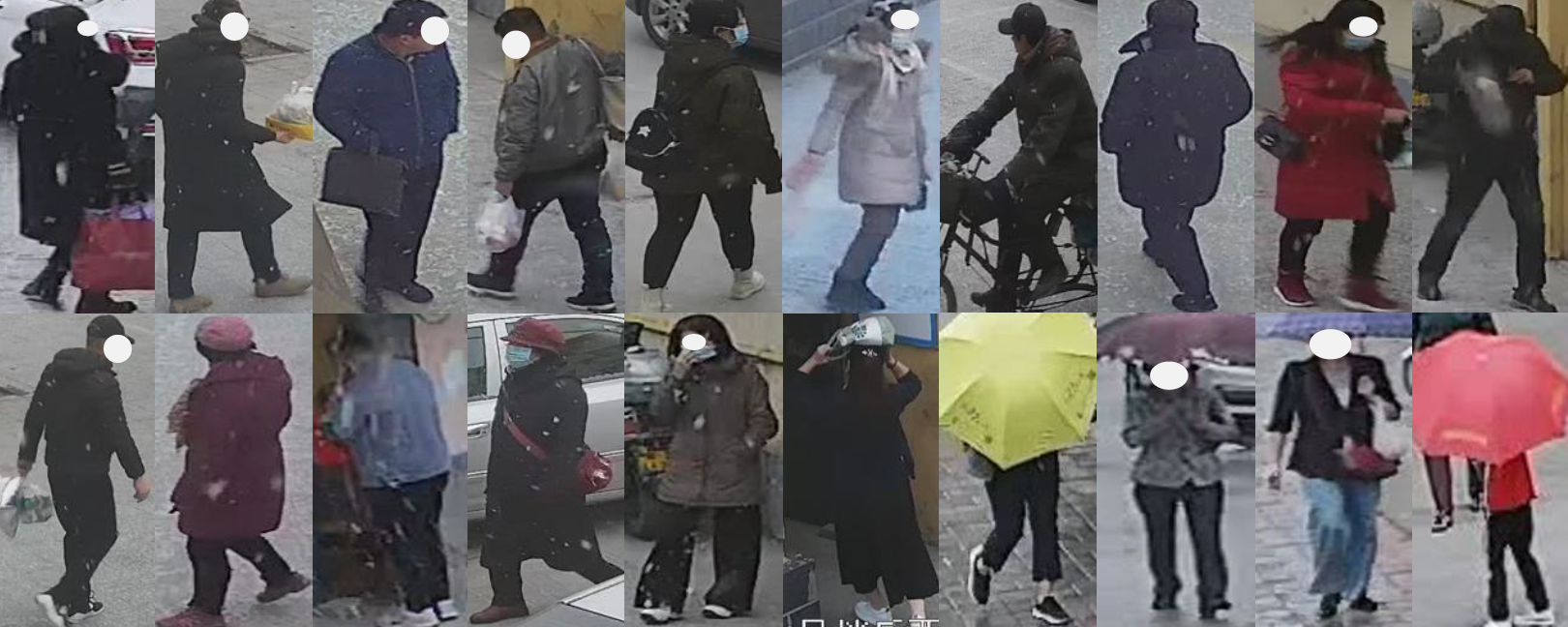}
		
	\caption{Samples collected in snow (bbox\#1-\#15) and rain (bbox\#16-\#20) weather from \name{}. Best viewed in color.}
	\label{fig:snow-and-rain-supp}
\end{figure*}

Figure~\ref{fig:snow-and-rain-supp} shows more samples with snow and rain, where we can see that snow and rain can cause noticeable appearance changes on both clothes and background.

\begin{figure*}[!h]
	\centering	
	\includegraphics[width=\linewidth]{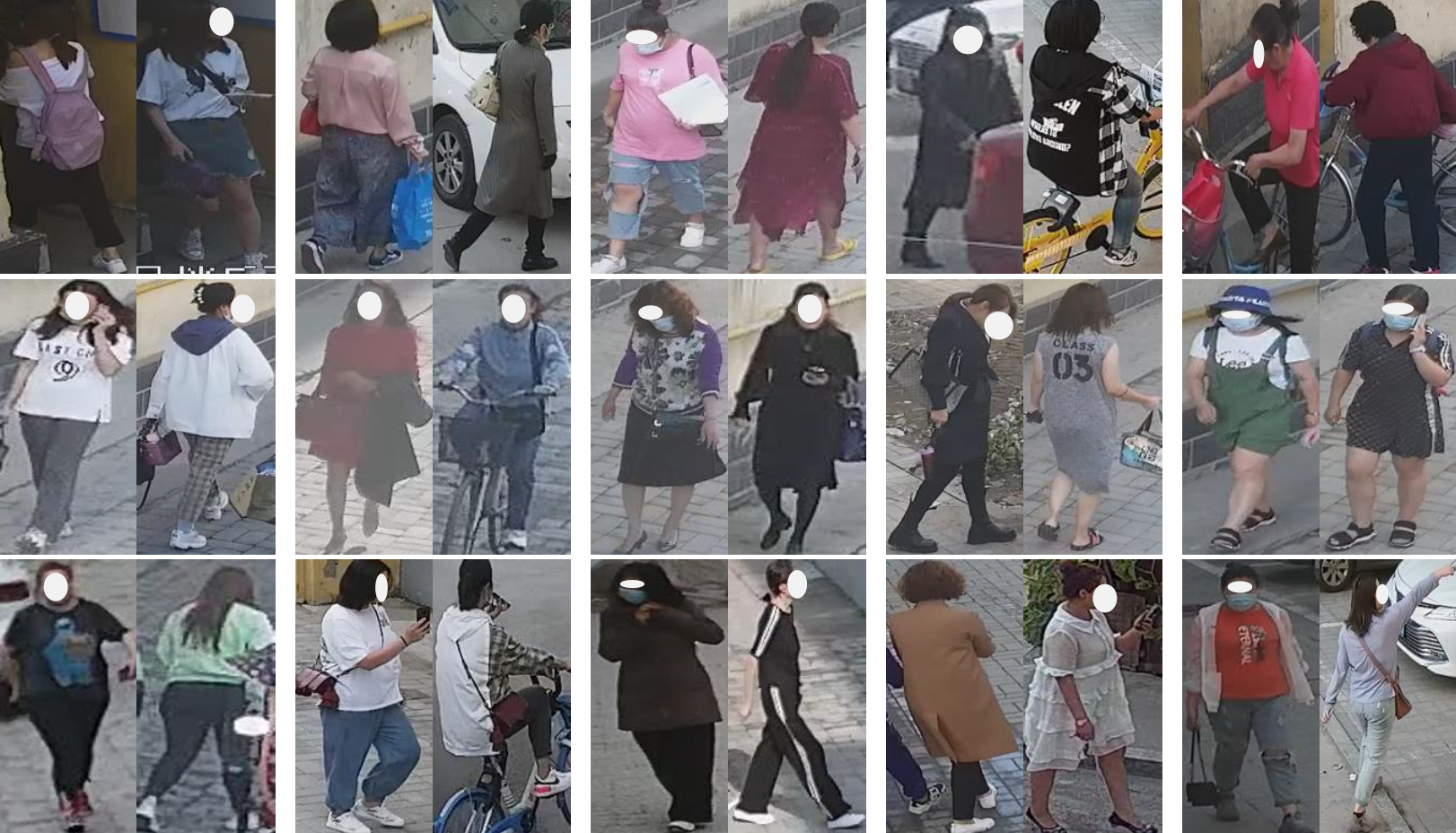}
	\caption{Sample pairs of specific persons with simultaneous clothes and hair style changes in the \name{} dataset. For each identity, only two different cases are selected randomly. Best viewed in color.}
	\label{fig:hair-style-supp}
\end{figure*}

Figure~\ref{fig:hair-style-supp} provides other random cases with simultaneous clothes and hair style changes.

\begin{figure*}[!h]
	\centering	
	\includegraphics[width=\linewidth]{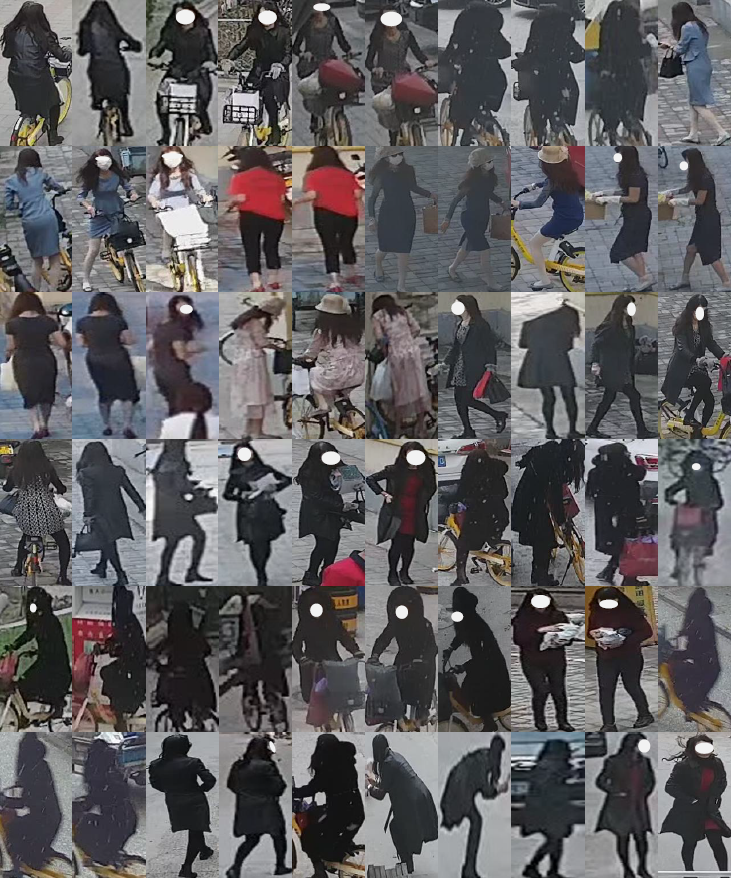}
	\caption{Images of a random person identity from \name{} with clothes change, as well as variation in behavior, pose, hair, illumination, weather, background, \etc. Best viewed in color.}
	\label{fig:p0318}
\end{figure*}

Figure~\ref{fig:p0318} demonstrates multiple bounding
boxes of an identity randomly selected from our \name{} dataset, where we can observe dramatic appearance
changes across weathers, seasons, \etc.

\end{document}